%% file: main.tex
\documentclass[runningheads]{llncs}

\PassOptionsToPackage{table}{xcolor}

\usepackage{eccv}

\usepackage{eccvabbrv}
\usepackage{graphicx}
\usepackage{booktabs}
\usepackage[accsupp]{axessibility}
\usepackage{algorithm}
\usepackage{multirow}
\usepackage{makecell}

\definecolor{calibgray}{gray}{0.5}
\definecolor{bestPink}{rgb}{0.980, 0.502, 0.447}
\definecolor{secondOrange}{rgb}{1.000, 0.784, 0.486}

\colorlet{colorFst}{bestPink!40}
\colorlet{colorSnd}{secondOrange!45}
\colorlet{colorTrd}{Goldenrod!20}

\newcommand{\best}[1]{\cellcolor{colorFst}#1}
\newcommand{\second}[1]{\cellcolor{colorSnd}#1}
\newcommand{\calib}[1]{\textcolor{calibgray}{#1}}

\usepackage{hyperref}
\usepackage{orcidlink}

\usepackage{hyperref}

\usepackage{orcidlink}

\begin{document}

\title{UniSim-SLAM: Feed-Forward SLAM with Unified
\texorpdfstring{$\mathrm{Sim}(3)$}{Sim(3)} Optimization}

\titlerunning{UniSim-SLAM}

\author{
Inha Lee\inst{}\orcidlink{0009-0003-8030-3131} \and
Dongjae Jeong\inst{}\orcidlink{0009-0008-8108-9102} \and
Junhee Lee\inst{}\orcidlink{0009-0001-4078-8811} \and
Kyungdon Joo\inst{}\orcidlink{0000-0002-3920-9608}
}

\authorrunning{I.~Lee et al.}

\institute{
lsan National Institute of Science and Technology, Ulsan, Korea \\
\email{\{epsilon8854,jeong\_dongjae,junhee98,kyungdon\}@unist.ac.kr}
}

\maketitle




\input{sec/0_abstract}
\input{sec/1_intro}
\input{sec/2_related}
\input{sec/3_method}

\input{sec/4_experiments}

\input{sec/5_conclusion}

\section*{Acknowledgements}
This work was supported by the Institute of Information \& Communications Technology Planning \& Evaluation (IITP) grant funded by the Korea government (MSIT) (No.RS-2020-II201336, Artificial Intelligence Graduate School Program (UNIST);
No.RS-2022-II220907, Development of AI Bots Collaboration Platform and Self-organizing;
No.RS-2026-25507551, Development of Egocentric Data Sensing and Spatial Immersive Experience Technology), and by the InnoCORE program of the Ministry of Science and ICT (25-InnoCORE-01).


%
%
\bibliographystyle{splncs04}
\bibliography{main}
\input{supple}
\end{document}

%% file: sec/0_abstract.tex
\begin{abstract}
Recent geometric foundation models enable feed-forward inference for SLAM, but their predictions are strongly dependent on the input view set, which leads to geometric inconsistencies and trajectory drift when results are chained over long sequences. Online deployment further exposes a trade-off between the low latency of two-view tracking and the constraint richness of multi-view inference. We introduce \texttt{UniSim-SLAM}, an integrated system that runs lightweight two-view keyframe tracking in the frontend and performs periodic multi-view submap refinement in the backend. To combine predictions defined in heterogeneous local coordinates with inconsistent scales, we formulate a unified multi-level factor graph on $Sim(3)$ that jointly optimizes global keyframe poses and submap poses. The graph integrates temporal view-to-view odometry edges, view-to-submap bridge edges with depth-statistics scale anchoring, and submap-to-submap tie and scale constraints to enforce consistent similarity relations across submaps. Experiments on TUM RGB-D and 7-Scenes show that \texttt{UniSim-SLAM} achieves state-of-the-art accuracy in the uncalibrated setting, reducing trajectory error by $38.5\% $ on TUM RGB-D and $45.9\%$ on 7-Scenes compared to prior best results. Project page: \url{https://vision3d-lab.github.io/unisim-slam/}.
  \keywords{Visual SLAM \and Feed Forward SLAM \and $Sim(3)$ Factor Graph }
\end{abstract}

%% file: sec/1_intro.tex
\section{Introduction}
\label{sec:introduction}




Recent progress in geometric foundation models is reshaping the design of visual SLAM systems.
For example, models such as DUSt3R~\cite{dust3r} and VGGT~\cite{vggt} have enabled feed-forward estimation of dense depth maps and relative camera poses, even from uncalibrated image sets.
As a result, recent SLAM systems~\cite{mast3rslam, vistaslam, vggtslam} have begun to apply feed-forward models to specific view subsets, such as image pairs or multi-view image clips, 
treating each inference output as a local reconstruction defined in its own local coordinate, which is subsequently aligned 
into a global coordinate system.
%

\begin{figure}
    \centering
    \includegraphics[width=0.98\linewidth]{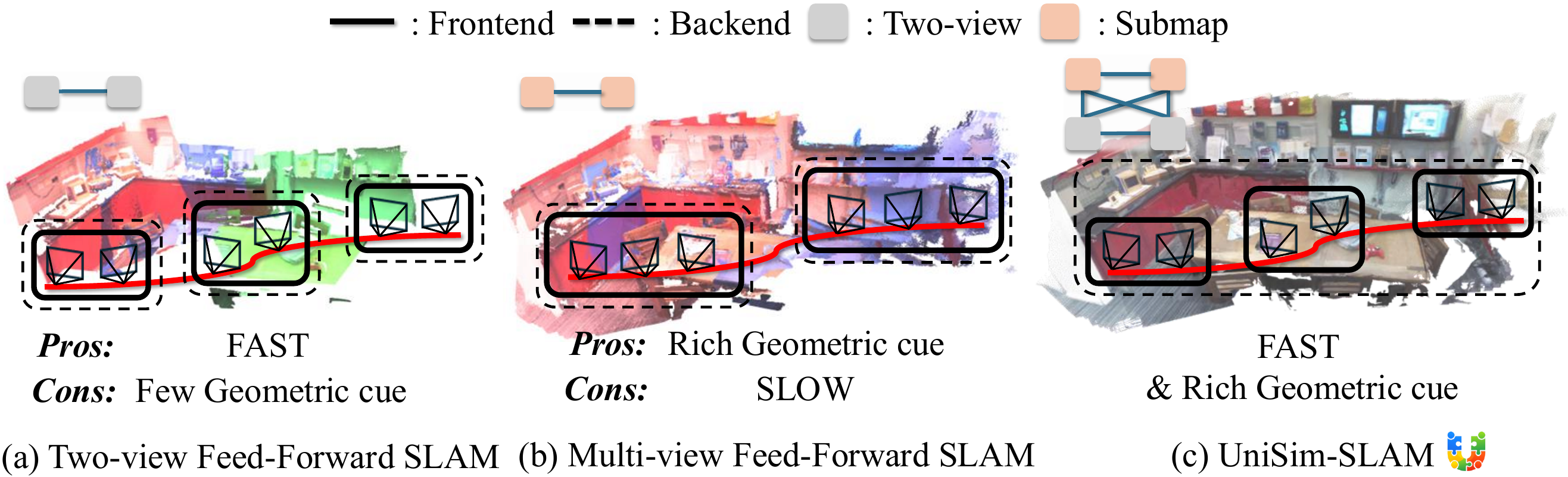}
    \caption{\textbf{Illustration of \texttt{UniSim-SLAM}.} Trade-off in feed-forward SLAM inference regimes. (a) Two-view inference enables low-latency tracking but provides limited geometric constraints. (b) Multi-view submap inference yields richer geometry but incurs higher latency. (c) \texttt{UniSim-SLAM} combines a two-view frontend with a multi-view submap backend and jointly optimizes both in a unified $Sim(3)$ factor graph.
    }
    \label{fig:teasure}
\end{figure}
One practical challenge in feed-forward SLAM is that geometric estimates are conditioned on the input view configuration. 
Even for the same image, the estimated scale and pose can vary depending on the co-visible views, 
leading to geometric inconsistencies when predictions are chained over long sequences.
%
Accordingly, the global trajectory must be constructed by aligning multiple configuration-dependent local reconstructions, each defined in its own coordinate frame.
For instance, two-view approaches align pairwise reconstructions using mutual constraints (\emph{e.g.}, VISTA-SLAM~\cite{vistaslam}), whereas multi-view methods register submaps by estimating transformations across overlapping regions (\emph{e.g.}, VGGT-SLAM~\cite{vggtslam}).

Beyond this configuration-dependent inconsistency, relying on a single inference regime (\emph{i.e.}, two-view or multi-view inference) further introduces a fundamental trade-off (see Fig.~\ref{fig:teasure}).
Multi-view inference leverages rich geometric constraints, but it requires accumulating a fixed number of frames (\emph{i.e.}, submap) before inference, making per-frame updates impractical.
%
Moreover, refinement based solely on submap relationships requires overlap between neighboring submaps to propagate corrections across the trajectory. 
In particular, when the overlap does not exist, geometric corrections cannot be effectively propagated, limiting global consistency.
In contrast, two-view inference is computationally lightweight and operates immediately upon receiving a new frame, resulting in low latency while naturally maintaining temporal connectivity. 
However, due to its limited geometric constraints by two-view inference, it is prone to drift accumulation over time, 
making it insufficient for maintaining long-term global consistency.
Therefore, relying on either inference regime alone is insufficient to achieve both low-latency tracking and long-term geometric consistency.


To address this trade-off, we revisit the architectural principles of classic SLAM systems.
Classic SLAM systems~\cite{orbslam, lsdslam,kimera,dso} successfully mitigate a similar {efficiency–consistency trade-off} by combining lightweight two-view odometry in the frontend with intermittent, globally consistent refinement in the backend. 
%
This paradigm ensures robust graph connectivity through the temporal consistency of two-view inference while simultaneously leveraging the rich geometric constraints of multi-view submaps.
However, existing feed-forward methods typically perform either pairwise two-view alignment~\cite{vistaslam} or submap-to-submap registration in isolation~\cite{vggtslam}.
%
Yet, naively combining two-view and multi-view predictions is non-trivial because their predictions reside in heterogeneous local coordinate systems with inconsistent scales and reference frames.
Consequently, a new factor graph formulation is required to jointly optimize these heterogeneous constraints within a single unified framework.

In this work, we propose \texttt{UniSim-SLAM}, a new feed-forward SLAM system that unifies two-view and multi-view inferences within a single $Sim(3)$ factor graph (see Fig.~\ref{fig:teasure}). 
%
Our key insight is that geometric predictions obtained from different feed-forward inference regimes can be interpreted as complementary constraints that reside in heterogeneous local coordinate systems, and therefore must be jointly optimized within a unified $Sim(3)$ factor graph.
Specifically, two-view inference provides lightweight and densely connected temporal constraints that support low-latency tracking, while multi-view inference produces geometrically consistent submaps that anchor the global structure of the scene.
UniSim-SLAM integrates these heterogeneous constraints through a unified multi-level factor graph defined on the $Sim(3)$ manifold, enabling consistent optimization across both frame-level and submap-level predictions.
By jointly optimizing two-view and multi-view constraints within this unified framework, \texttt{UniSim-SLAM} resolves the efficiency-consistency trade-off in feed-forward SLAM, achieving both low-latency tracking and long-term geometric stability. 
%
{We evaluate \texttt{UniSim-SLAM} on standard SLAM benchmarks~\cite{tumrgbd,7scenes}, demonstrating state-of-the-art performance.} 
The main contributions are summarized as follows:
\begin{itemize}
\item We propose \texttt{UniSim-SLAM}, a feed-forward SLAM system that jointly optimizes heterogeneous predictions from two-view and multi-view inference regimes within a unified $Sim(3)$ optimization framework.
\item We introduce a unified multi-level factor graph that connects frames and submaps through three complementary constraints
(view-to-view, submap-to-view, and submap-to-submap edges), enabling scale-consistent optimization and drift correction even when submaps do not directly overlap.
\item We demonstrate that integrating temporally dense two-view constraints with geometrically rich multi-view submaps enables
robust correction propagation across long trajectories, achieving state-of-the-art performance on standard SLAM benchmarks.
\end{itemize}

%% file: sec/2_related.tex
\section{Related Work}
\label{sec:related}

\subsection{Visual SLAM}
Visual SLAM can be broadly categorized into two paradigms, feature-based methods and direct methods.
Feature-based methods such as ORB-SLAM~\cite{orbslam, orbslam2, orbslam3} and Kimera~\cite{kimera} follow the SfM pipeline~\cite{sfm, incrementalsfm, sfmrevisited}, leveraging keypoint matching for triangulation and PnP~\cite{pnp, epnp}. 
In contrast, direct methods~\cite{lsdslam, dso} optimize camera poses by minimizing photometric residuals, typically alongside per-frame depth estimation.
Despite different frontends, both typically rely on an optimization backend, most commonly bundle adjustment~\cite{ba}.
They are sensitive to camera calibration and challenging visual conditions, often producing only sparse or semi-dense maps.


To overcome these limitations, learning-based SLAM integrates deep neural networks into the frontend or the scene representation to improve robustness and densify mapping.
DeepFactors~\cite{deepfactors} and DeepV2D~\cite{deepv2d} focus on learned depth and pose with multi-view consistency, whereas DROID-SLAM~\cite{droidslam} and DPV-SLAM~\cite{dpvslam} couple learned correspondences with differentiable bundle adjustment for iterative refinement.
Neural implicit mapping jointly optimizes camera poses with an implicit scene representation online.
For example, iMAP~\cite{imap} and NICER-SLAM~\cite{nicerslam} use MLP-based implicit maps for online reconstruction, while GlORIE-SLAM~\cite{glorieslam} adopts a deformable neural point-cloud representation with loop closure and online global BA for improved global consistency.
More recently, 3D Gaussian Splatting~\cite{3dgs} has enabled efficient differentiable dense SLAM. 
3DGS-based methods, including Gaussian Splatting SLAM~\cite{gaussiansplattingslam} and GS-SLAM~\cite{gsslam}, jointly optimize poses and Gaussian primitives.

Nevertheless, most learning-based and neural mapping approaches still rely on accurate intrinsics and remain computationally heavy for real-time use.

\subsection{Feed-Forward Visual SLAM}
Recent progress in large-scale self-supervised representation learning and geometry estimation (\eg, DINOv2~\cite{dinov2}, Depth Anything~\cite{depthanything, depthanythingv2, depthanythingv3}, FoundationStereo~\cite{foundationstereo}) has enabled feed-forward 3D geometric foundation models that predict geometry and camera parameters in a feed-forward manner, reducing reliance on handcrafted correspondences and heavy per-sequence optimization.
Based on these advances, pairwise feed-forward models~\cite{dust3r, mast3r} regress 3D structure and dense correspondences from uncalibrated image pairs, and can be scaled to unconstrained collections via retrieval-based view graphs and global alignment~\cite{mast3rsfm}.
Complementary directions include direct relative pose regression~\cite{reloc3r} and feed-forward prediction of 3D Gaussian primitives without intrinsics~\cite{splatt3r}.

Beyond pairwise settings, these models also support multi-view and long-horizon inference.
Memory or state-based models~\cite{spann3r, cut3r} enable incremental reconstruction in a global frame, while long-term tracking provides a robust correspondence backbone~\cite{cotracker}.
Recent multi-view foundation models further infer dense geometry jointly with camera parameters from one to many views in a single pass (\eg, VGGT~\cite{vggt}, $\pi^3$~\cite{pi3}, MapAnything~\cite{mapanything}), with VGGT-Long~\cite{vggtlong} improving scalability to long RGB streams via chunk-wise reconstruction and lightweight loop-closure optimization.

Leveraging these models, feed-forward visual SLAM methods have emerged and can be broadly grouped into two-view and multi-view pipelines.
Two-view feed-forward SLAM conducts pairwise predictions and lightweight global optimization.
For example, MASt3R-SLAM~\cite{mast3rslam} leverages MASt3R as a two-view 3D reconstruction and matching prior and builds a real-time dense monocular SLAM system with pointmap-based matching, local fusion, loop closure, and second-order global optimization.
Similarly, ViSTA-SLAM~\cite{vistaslam} proposes a symmetric two-view association frontend that regresses local point maps and relative pose from two RGB images, mitigating drift via $Sim(3)$ pose-graph optimization with loop closure in the backend.
Multi-view feed-forward SLAM focuses on incrementally aligning and globally optimizing submaps produced by multi-view reconstruction backbones.
VGGT-SLAM~\cite{vggtslam} incrementally builds VGGT submaps and globally aligns them by optimizing 15-DoF projective transformations on the $SL(4)$ manifold to handle ambiguity under uncalibrated cameras.


Most feed-forward SLAM systems exploit either two-view temporal constraints or multi-view submap constraints in isolation.
In contrast, \texttt{UniSim-SLAM} jointly utilizes both within a unified multi-level factor graph.

%% file: sec/3_method.tex
\section{Method}    \label{sec:method}
\subsection{Overview}
\label{sec:method_overview}
\begin{figure}[t]
    \centering
    \includegraphics[width=0.98\linewidth]{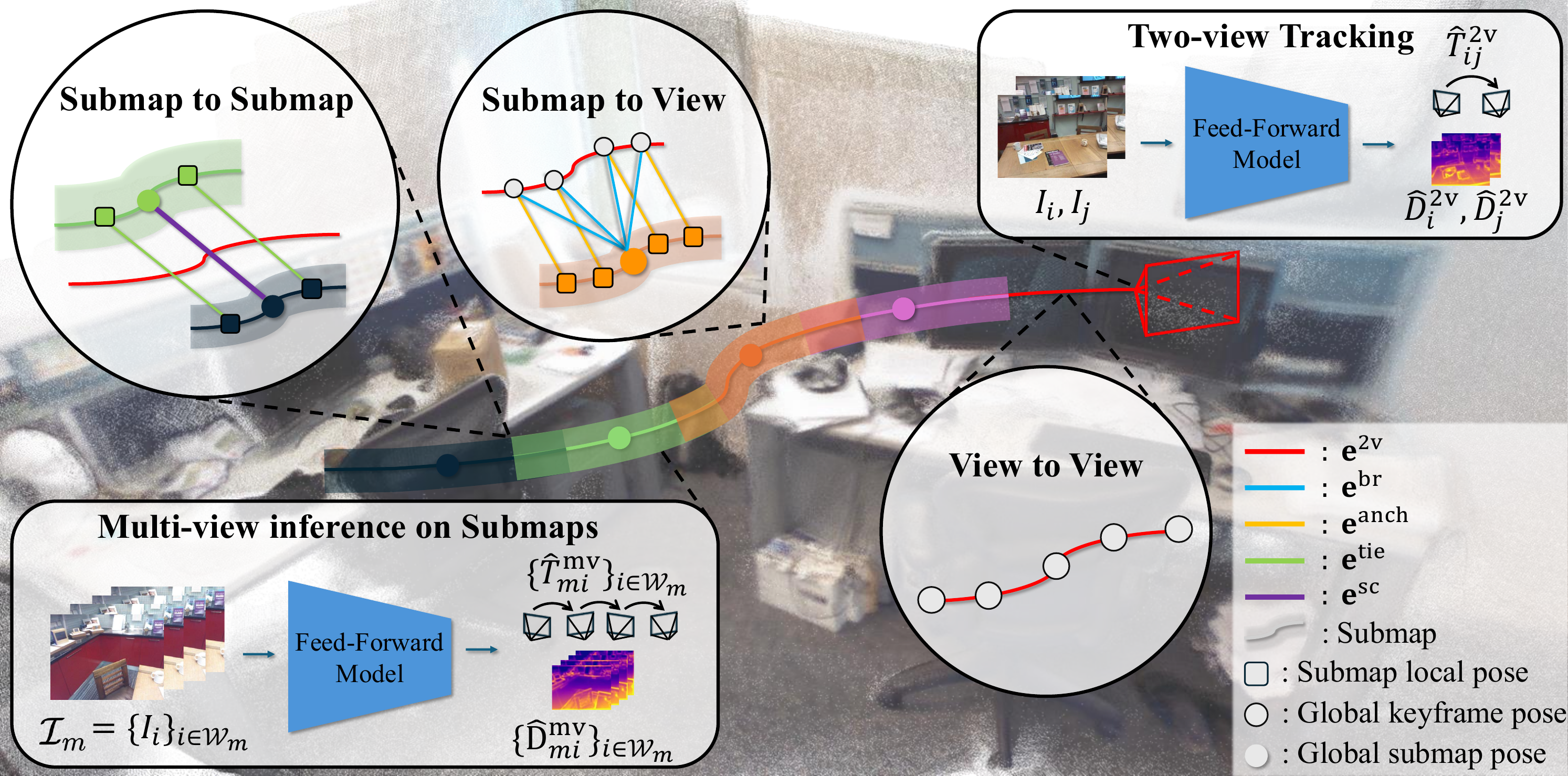}
    \caption{\textbf{Overall framework of \texttt{UniSim-SLAM}.}
    A feed-forward model produces two-view pose and depth for tracking and multi-view submap predictions for local geometry. These predictions are integrated into a unified $Sim(3)$ factor graph with multiple edge types, jointly optimizing global keyframes and submap poses.
    }
    \label{fig:overview}
\end{figure}

\texttt{UniSim-SLAM} revisits the classical SLAM paradigm to address the trade-off between inference latency and geometric constraint richness in feed-forward SLAM. 
Our system combines a lightweight two-view frontend for low-latency tracking with a constraint-rich multi-view backend for drift correction.
%
However, two-view and multi-view inferences are defined in different local coordinate systems and exhibit inconsistent scale and reference frames. 
%
To coherently integrate these heterogeneous predictions, we formulate a unified pose graph on the $Sim(3)$ manifold, which naturally accommodates rotation, translation, and scale ambiguity (see Fig.~\ref{fig:overview}).

Given an input image stream, the frontend samples keyframes and incrementally estimates global poses using pairwise two-view predictions. In parallel, the backend periodically constructs multi-view submaps over contiguous keyframe windows, producing submap-local pose estimates.
The optimization variables of the unified factor graph consist of global keyframe poses $\{T_i\}$ and submap poses $\{S_m\}$, both of which are jointly refined within a unified $Sim(3)$ optimization framework.
%
 Details of the frontend and backend are described in Secs.~\ref{sec:method_frontend} and~\ref{sec:method_backend}, followed by the unified factor graph formulation in Sec.~\ref{sec:pose_graph}.

\subsection{Two-View Tracking on Keyframes} \label{sec:method_frontend}
The frontend plays a central role in ensuring low latency and temporal consistency. 
For each temporally consecutive keyframe pair $(I_i, I_j)$ with $j = i+1$,
we feed the image pair into a feed-forward model (\emph{e.g.}, VGGT~\cite{vggt}, STA~\cite{vistaslam}) to obtain two-view estimations: 
\begin{equation}
\{\hat D^{\text{2v}}_i, \hat D^{\text{2v}}_j, \hat{T}^{\text{2v}}_{ij}\} = f_\text{2v}(I_i, I_j),
\end{equation}
where $\hat D^{\text{2v}}_i$ and $\hat D^{\text{2v}}_j$ denote the predicted depth maps, and $\hat{T}^{\mathrm{2v}}_{ij} \in Sim(3)$ represents the relative transformation from keyframe $I_i$ to $I_j$ with its scale component initialized to 1.
%
The predicted depth maps are later used to estimate scale anchors for submap integration.

To construct a global trajectory from these relative measurements, we first define a global reference frame.
The first keyframe $I_0$ defines the origin of the global coordinate system.
Then, global poses are initialized online by sequentially composing the two-view relative transformations:
\begin{equation}
T_j = T_i \hat{T}^{\mathrm{2v}}_{ij},
\end{equation}
This sequential initialization implicitly defines the temporal edges of the pose graph and maintains its connectivity even when submaps do not overlap.
The resulting trajectory serves as the initial estimate, which is later refined by multi-view submap constraints in the backend.

\subsection{Multi-view Submap Integration} \label{sec:method_backend}

The goal of the backend is to correct the accumulated drift in the global keyframe poses $\{T_i\}$.
To achieve this, we periodically construct multi-view submaps that introduce geometrically rich constraints into the global pose graph. 
Once a sufficient number of consecutive keyframes are accumulated, we construct the $m$-th submap over a contiguous window of keyframes indexed by $\mathcal{W}_m \subset \{1,\dots,N\}$.
The corresponding keyframe sequence is denoted by $\mathcal{I}_m = \{I_i\}_{i \in \mathcal{W}_m}$.
We then apply a feed-forward multi-view model to obtain:
\begin{equation}
   \{ \hat{D}^{\mathrm{mv}}_{mi}, \hat{T}^{\mathrm{mv}}_{mi} \}_{i \in \mathcal{W}_m}
= f_{\mathrm{mv}}(\mathcal{I}_m), 
\end{equation}
where $\hat{D}^{\mathrm{mv}}_{mi}$ denotes the predicted depth map for keyframe $I_i$ within the $m$-th submap, and $\hat{T}^{\mathrm{mv}}_{mi} \in Sim(3)$ represents the submap-local pose of $I_i$, with its scale component initialized to 1.
%
Although the same geometric foundation model is employed as in the two-view frontend, we denote it by $f_{\mathrm{mv}}$ to highlight its operation on multi-view inputs and its asynchronous execution in the backend.

\begin{figure}[t]
    \centering
    \includegraphics[width=0.95\linewidth]{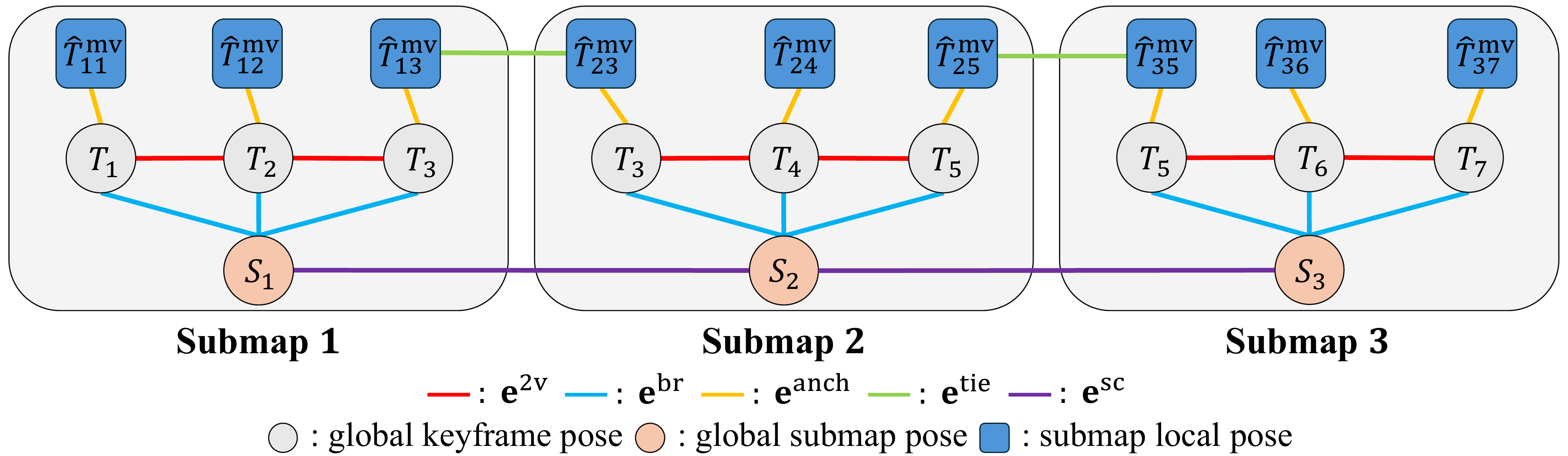}
    \caption{\textbf{Representation of multi-level factor graph.} 
    }
    \label{fig:method_detail}
\end{figure}

During inference, each submap is reconstructed in its own local coordinate frame, with the center keyframe in $\mathcal{W}_m$ as the submap origin. 
To integrate this reconstruction into the global trajectory, 
we introduce a submap pose $S_m \in Sim(3)$ that transforms the m-th submap coordinate frame into the global frame:
\begin{equation}
    T_i \approx S_m \hat T^\text{mv}_{mi}.
\end{equation}
The optimization therefore jointly estimates  the global keyframe poses $\{T_i\}$ and the submap poses $\{S_m\}$.
%
Because multi-view predictions are view-set dependent, the same keyframe can appear with different scales and poses across submaps. Hence, submap-local poses cannot be treated as globally consistent measurements without explicitly introducing $\{S_m\}$.
To successfully perform this joint optimization,
we initialize the submap pose $S_m$ before joint optimization. Let $I_i$ denote the origin keyframe of submap $m$, such that $\hat{T}^{\mathrm{mv}}_{mi} = \mathbf{I}$.
Let the corresponding global pose obtained from two-view inference be:
\begin{equation}
T_i =
\begin{bmatrix}
s_i R_i & t_i \\
\mathbf{0}^\top & 1
\end{bmatrix},    
\end{equation}
where $R_i \in \mathrm{SO}(3)$, $t_i \in \mathbb{R}^3$, and $s_i \in \mathbb{R}^+$.
To resolve the relative scale ambiguity between two-view and multi-view predictions, we estimate a relative scale using depth statistics $
s^{\mathrm{rel}}_{mi} = \mathrm{median} ({\hat{D}^{\mathrm{mv}}_{mi}} /  {\hat{D}^{\mathrm{2v}}_{i}})$.
Since the multi-view prediction is expressed with unit scale, the initial submap scale is set as $s_m^{(0)} = {s_i}/{s^{\mathrm{rel}}_{mi}}$.
The submap pose is then initialized in $Sim(3)$ form as: 
\begin{equation}
S_m^{(0)} =
\begin{bmatrix}
s_m^{(0)} R_i & \dfrac{t_i}{s^{\mathrm{rel}}_{mi}} \\
\mathbf{0}^\top & 1
\end{bmatrix}.
\end{equation}
This initialization ensures that the submap origin aligns with the global pose of $I_i$ while maintaining a consistent similarity transformation structure. 
The translation is scaled by the same relative factor to preserve the similarity transformation between the global and submap frames.
It provides a scale-consistent embedding of the submap into the global trajectory, after which both $T_i$ and $S_m$ are jointly refined in the unified optimization.

However, when two-view and multi-view constraints are simultaneously imposed, multiple $Sim(3)$ relationships coexist within a submap. Each keyframe is associated with a global pose $T_i$, while the submap is associated with its own $Sim(3)$ variable $S_m$. Consequently, enforcing a single relative scale constraint at initialization is insufficient to guarantee global consistency. 
This observation motivates the multi-level factor graph formulation, which explicitly models heterogeneous $Sim(3)$ constraints while jointly optimizing $\{T_i\}$ and $\{S_m\}$.

\subsection{Unified $Sim(3)$ Pose Graph}  \label{sec:pose_graph}

As discussed in Sec.~\ref{sec:method_backend}, two-view and multi-view feed-forward inferences produce $Sim(3)$ predictions in different coordinate systems with independent scale. 
Directly enforcing these heterogeneous constraints in a conventional pose graph may lead to incompatible similarity relations and unstable optimization.
To overcome this structural inconsistency, we formulate a unified multi-level pose graph(Fig.~\ref{fig:method_detail}) directly on the $Sim(3)$ manifold.
Unlike prior pipelines that optimize pairwise or submap constraints in isolation, our formulation jointly models all heterogeneous $Sim(3)$ relations in a single similarity-consistent graph.
%
The unified graph $\mathcal{G} = (\mathcal{V}, \mathcal{E})$ consists of global pose nodes and submap pose nodes:
\begin{equation}
\mathcal{V} = \mathcal{V}_T \cup \mathcal{V}_S, 
\quad 
\mathcal{V}_T = \{T_i \in Sim(3)\}, 
\quad 
\mathcal{V}_S = \{S_m \in Sim(3)\}.
\end{equation}
The edge set is decomposed into three hierarchical groups:
\begin{equation}
\mathcal{E}
=
\mathcal{E}^{\mathrm{temp}}
\;\cup\;
\mathcal{E}^{\mathrm{v2s}}
\;\cup\;
\mathcal{E}^{\mathrm{s2s}},
\end{equation}
encoding geometric relations across temporal, cross-level, and inter-submap structures. Specifically, 
\emph{temporal edges} $\mathcal{E}^{\mathrm{temp}}$ connect consecutive global poses to ensure connectivity. 
\emph{View-to-submap edges} $\mathcal{E}^{\mathrm{v2s}}$ link global poses $T_i$ with submap poses $S_m$ to align local predictions and constrain relative scales. Finally, 
\emph{submap-to-submap edges} $\mathcal{E}^{\mathrm{s2s}}$ directly connect overlapping submaps to enforce similarity consistency and prevent scale drift.
%

%
\vspace{1mm} \noindent \textbf{View-to-View Edges.} \ 
The temporal edge set is defined as $\mathcal{E}^{\mathrm{temp}} = \{ (i,j) \mid j = i+1 \}$,
where each pair represents a temporally consecutive keyframe pair.
For each $(i,j) \in \mathcal{E}^{\mathrm{temp}}$, the two-view feed-forward model provides a relative $Sim(3)$ measurement $\hat{T}^{\mathrm{2v}}_{ij}$.
We impose the temporal consistency constraint $T_i^{-1} T_j \approx \hat{T}^{\mathrm{2v}}_{ij}$, which enforces agreement between the composed global poses and the pairwise feed-forward prediction.

Beyond local consistency, these edges form a globally connected temporal backbone. 
This connectivity is crucial in the unified multi-level graph: even when submaps are sparse or non-overlapping,
temporal edges allow corrections from higher-level constraints to propagate across the trajectory.
The corresponding residual in the Lie algebra $\mathfrak{sim}(3)$ is defined as
\begin{equation}
\mathbf{e}^{\mathrm{2v}}_{ij}
=
\log
\!\left(
({\hat{T}^{\mathrm{2v}}_{ij}})^{-1}
(T_i^{-1} T_j)
\right).
\end{equation}



\vspace{1mm} \noindent \textbf{View-to-Submap Edges.} \ 
The view-to-submap edge set is defined as $\mathcal{E}^{\mathrm{v2s}} = \{ (m,i) \mid i \in \mathcal{W}_m \}$, where each element connects a global pose node $T_i$ with its corresponding submap pose node $S_m$.
%
For each $(m,i) \in \mathcal{E}^{\mathrm{v2s}}$, submap $m$ contains the submap-local pose $\hat{T}^{\mathrm{mv}}_{mi}$.
Each view-to-submap edge introduces two complementary residual terms: a pose alignment constraint and a scale consistency constraint.

\noindent \textit{Pose alignment.} \ 
We enforce the $Sim(3)$ consistency relation $T_i \approx S_m \hat{T}^{\mathrm{mv}}_{mi}$, which aligns the submap-local prediction with the global trajectory.
The corresponding bridge residual in $\mathfrak{sim}(3)$ is defined as
\begin{equation}
\mathbf{e}^{\mathrm{br}}_{mi}
=
\log\!\left(
(\hat{T}^{\mathrm{mv}}_{mi})^{-1}
S_m^{-1} T_i
\right).
\end{equation}
Bridge edges align submap-local predictions with the global trajectory, correcting drift within $\mathcal{W}_m$.
When keyframes are shared across submaps, this coupling implicitly aligns overlapping submaps through their common global pose nodes.
However, pose alignment alone does not constrain the relative scale between the global trajectory and the submap coordinate frame.

\noindent \textit{Scale anchoring.} \ 
In addition to pose alignment, each view–submap edge also constrains relative scale using per-view depth statistics. 
Let $\hat{D}^{\mathrm{2v}}_i$ and $\hat{D}^{\mathrm{mv}}_{mi}$
denote the predicted two-view and multi-view depths.
The scale residual is defined as
\begin{equation}
\mathbf{e}^{\mathrm{anch}}_{mi} = \log s_i - \log s_m - 
\log\!\left( \mathrm{median} ( {\hat{D}^{\mathrm{mv}}_{mi}} / {\hat{D}^{\mathrm{2v}}_{i}} ) \right).   
\end{equation}
This additional scale term stabilizes the relative scale between the global trajectory and the submap coordinate frame, preventing scale inconsistency from propagating across submaps.

\vspace{1mm}\noindent\textbf{Submap-to-Submap Edges.} \ 
While view-to-submap edges enable indirect alignment between submaps via global view nodes, this coupling can be compensated by shifting intermediate view nodes.
To enforce strict consistency between submap coordinate frames, we introduce direct submap-to-submap edges for overlapping submaps.
The submap-to-submap edge set is defined as $\mathcal{E}^{\mathrm{s2s}} =\{(m,n) \mid \mathcal{W}_m \cap \mathcal{W}_n \neq \emptyset \}$, where an edge is introduced between submaps whose keyframe windows overlap.
Let $\mathcal{V}_{mn} = \mathcal{W}_m \cap \mathcal{W}_n$ denote the shared keyframes between submaps $m$ and $n$.
Each submap-to-submap edge introduces two residual terms: a pose consistency constraint (tie) and a scale consistency constraint.

\noindent \textit{Pose consistency (tie).} 
For each shared keyframe $i \in \mathcal{V}_{mn}$, multi-view inference yields independent local pose predictions $\hat{T}^{\mathrm{mv}}_{mi}$ and
$\hat{T}^{\mathrm{mv}}_{ni}$ within submaps $m$ and $n$, respectively. 
Geometric consistency requires that transforming the shared view from both submap coordinate frames yields the same global pose, $S_m \hat{T}^{\mathrm{mv}}_{mi} \approx S_n \hat{T}^{\mathrm{mv}}_{ni}$.
We encode this requirement with the tie residual
\begin{equation}
\mathbf{e}^{\mathrm{tie}}_{mni}
=
\log
\left(
(S_m \hat{T}^{\mathrm{mv}}_{mi})^{-1}
(S_n \hat{T}^{\mathrm{mv}}_{ni})
\right).    
\end{equation}

We also enforce scale consistency between overlapping submaps using dense depth statistics.
For each shared keyframe $i \in \mathcal{V}_{mn}$, we estimate a per-view relative scale $\hat{s}_{mn,i}
= \mathrm{median} ({\hat{D}^{\mathrm{mv}}_{mi}} / {\hat{D}^{\mathrm{mv}}_{ni}} )$,
where $\hat{D}^{\mathrm{mv}}_{mi}$ and $\hat{D}^{\mathrm{mv}}_{ni}$ denote the multi-view depth predictions for keyframe $i$ obtained within submaps $m$ and $n$, respectively.
We aggregate these estimates across shared views as
$\hat{s}_{mn}
=
\mathrm{median}_{i \in \mathcal{V}_{mn}}
\hat{s}_{mn,i}$.
The corresponding scale residual is defined as
\begin{equation}
\mathbf{e}^{\mathrm{sc}}_{mn} = \log s_n -  \log s_m - \log \hat{s}_{mn}.
\end{equation}
%
Using statistics over pixel-aligned depth predictions from shared keyframes provides a robust estimate of the relative scale between submaps.
%
These tie and scale constraints align the coordinate frames and relative scales of overlapping submaps, preventing drift between independently estimated submaps and ensuring that the multi-level graph remains globally consistent.
This direct constraint prevents degenerate solutions where submap alignment is satisfied through compensating changes in intermediate view poses.

\vspace{1mm}\noindent\textbf{Optimization.} \ 
We jointly optimize the global view poses $\{T_i\}$ and submap poses
$\{S_m\}$ by minimizing the residuals introduced by the edges in the
unified $Sim(3)$ graph. The optimization variables lie on the $Sim(3)$
manifold and are solved using nonlinear least squares.
The overall objective is defined as
\begin{align}
\underset{{\{T_i\},\{S_m\}}}{\arg\min}
\;&
\sum_{(i,j)\in \mathcal{E}^{\mathrm{temp}}}
\rho
\!\left(
\left\|
\mathbf{e}^{2v}_{ij}
\right\|
\right)
+
\sum_{(m,i)\in \mathcal{E}^{\mathrm{v2s}}}
\Big(
\rho
\!\left(
\left\|
\mathbf{e}^{\mathrm{br}}_{mi}
\right\|
\right)
+
\rho
\!\left(
\left\|
\mathbf{e}^{\mathrm{anch}}_{mi}
\right\|
\right)
\Big)
\\ \nonumber
+
&
\sum_{(m,n)\in \mathcal{E}^{\mathrm{s2s}}}
\Big(
\sum_{i\in\mathcal{V}_{mn}}
\rho
\!\left(
\left\|
\mathbf{e}^{\mathrm{tie}}_{mni}
\right\|
\right)
+
\rho
\!\left(
\left\|
\mathbf{e}^{\mathrm{sc}}_{mn}
\right\|
\right)
\Big),
\end{align}
where $\rho(\cdot)$ denotes the Huber loss function. 
The optimization is solved using nonlinear least squares with the Levenberg--Marquardt algorithm on the $Sim(3)$ manifold via the Lie algebra $\mathfrak{sim}(3)$.
Through this unified optimization, geometric corrections introduced by higher-level submap constraints propagate consistently across the entire trajectory.


\vspace{1mm}\noindent\textbf{Loop Closure.} \ 
{To correct long-term drift, we incorporate a loop closure module following prior works~\cite{vggtslam}.
Loop candidates are detected via global image retrieval and geometric verification between keyframes.
Once a loop is detected, we construct a \emph{joint loop submap} that aggregates two multi-view sets centered at the query keyframe and the matched keyframe.
The resulting multi-view predictions introduce additional constraints by inserting the loop submap into the unified $Sim(3)$ pose graph. Details are provided in the supplementary material.}


%% file: sec/4_experiments.tex
\section{Experiment}
\label{sec:experiment}

\subsection{Experimental Setup}
 We evaluate \texttt{UniSim-SLAM} on two standard RGB SLAM benchmarks: TUM RGB-D~\cite{tumrgbd} and 7-Scenes~\cite{7scenes}. For 7-Scenes, we utilize the refined ground-truth poses provided by Brachmann et al.~\cite{brachmann2021limits}. To evaluate tracking performance, we measure the Root Mean Square Error (RMSE) of the Absolute Trajectory Error (ATE) following $Sim(3)$ alignment, computed via the evo toolkit~\cite{evo}. Furthermore, to assess dense 3D reconstruction quality, we report Accuracy, Completion, and Chamfer Distance on the 7-Scenes dataset.

We compare \texttt{UniSim-SLAM} against recent learning-based SLAM systems, including two-view feed-forward approaches (MASt3R-SLAM \cite{mast3rslam}, ViSTA-SLAM \cite{vistaslam}), multi-view approaches (VGGT-SLAM \cite{vggtslam}), and regression-based methods (CUT3R \cite{cut3r}, SLAM3R \cite{slam3r}). 
Calibration-based methods~\cite{orbslam3, deepv2d, deepfactors, dpvslam, goslam} assuming known intrinsics are also included. 
For TUM RGB-D, we report baseline results as reported in prior works~\cite{mast3rslam,vggtslam,vistaslam}. 

\vspace{1mm}\noindent \textbf{Implementation Details.} \
For 7-Scenes, we use a keyframe selection stride of 5 for all uncalibrated methods. 
For TUM RGB-D, we follow the evaluation protocol of~\cite{vistaslam} and use a stride of 3. For backend optimization, \texttt{UniSim-SLAM} constructs submaps of size 16 with an overlap of 2 frames. Additional implementation details are provided in the supplementary material.

\begin{table*}[t]
    \centering
    \caption{\textbf{Quantitative trajectory results on TUM RGB-D.} We report ATE RMSE $[m]$ $\downarrow$) for each sequence under calibrated (Calib.) and uncalibrated (Uncalib.) methods. ‘$\times$’ denotes failure to produce a valid trajectory.}
    \label{tab:ate_tumrgbd}
    \vspace{-2mm}
    \small
    \resizebox{1.0\linewidth}{!}{
    \setlength{\tabcolsep}{8pt} 

    \begin{tabular}{ll|ccccccccc|c}
        \toprule
        & Method & 360 & desk & desk2 & floor & plant & room & rpy & teddy & xyz & Avg \\
        \midrule
        
        \cellcolor{gray!15} & \calib{ORB-SLAM3~\cite{orbslam3}}    & \calib{$\times$} & \calib{0.017} & \calib{0.210} & \calib{$\times$} & \calib{0.034} & \calib{$\times$} & \calib{$\times$} & \calib{$\times$} & \calib{\textbf{0.009}} & \calib{N/A} \\
        \cellcolor{gray!15} & \calib{DeepV2D~\cite{deepv2d}}     & \calib{0.243} & \calib{0.166} & \calib{0.379} & \calib{1.653} & \calib{0.203} & \calib{0.246} & \calib{0.105} & \calib{0.316} & \calib{0.064} & \calib{0.375} \\
        \cellcolor{gray!15} & \calib{DeepFactors~\cite{deepfactors}} & \calib{0.159} & \calib{0.170} & \calib{0.253} & \calib{0.169} & \calib{0.305} & \calib{0.364} & \calib{0.043} & \calib{0.601} & \calib{0.035} & \calib{0.233} \\
        \cellcolor{gray!15} & \calib{DPV-SLAM~\cite{dpvslam}}    & \calib{0.112} & \calib{0.018} & \calib{0.029} & \calib{0.057} & \calib{0.021} & \calib{0.330} & \calib{0.030} & \calib{0.084} & \calib{0.010} & \calib{0.076} \\
        \cellcolor{gray!15} & \calib{DPV-SLAM++~\cite{dpvslam}}  & \calib{0.132} & \calib{0.018} & \calib{0.029} & \calib{0.050} & \calib{0.022} & \calib{0.096} & \calib{0.032} & \calib{0.098} & \calib{0.010} & \calib{0.054} \\
        \cellcolor{gray!15}& \calib{GO-SLAM~\cite{goslam}}     & \calib{0.089} & \calib{\textbf{0.016}} & \calib{0.028} & \calib{0.025} & \calib{0.026} & \calib{0.052} & \calib{\textbf{0.019}} & \calib{0.048} & \calib{0.010} & \calib{0.035} \\
        \cellcolor{gray!15} & \calib{DROID-SLAM~\cite{droidslam}}  & \calib{0.111} & \calib{0.018} & \calib{0.042} & \calib{\textbf{0.021}} & \calib{\textbf{0.016}} & \calib{\textbf{0.049}} & \calib{0.026} & \calib{0.048} & \calib{0.012} & \calib{0.038} \\
        \multirow{-8}{*}{\cellcolor{gray!15}\rotatebox{90}{\textit{Calib.}}} & \calib{MASt3R-SLAM~\cite{mast3rslam}} & \calib{\textbf{0.049}} & \calib{\textbf{0.016}} & \calib{\textbf{0.024}} & \calib{0.025} & \calib{0.020} & \calib{0.061} & \calib{0.027} & \calib{\textbf{0.041}} & \calib{\textbf{0.009}} & \calib{\textbf{0.030}} \\
        
        \midrule 
        \midrule
        
        \cellcolor{gray!15} & CUT3R~\cite{cut3r} & 0.174 & 0.592 & 0.546 & 0.662 & 0.467 & 0.911 & 0.051 & 0.845  & 0.129  & 0.486 \\
        \cellcolor{gray!15} & SLAM3R~\cite{slam3r} & 0.211 & 0.861 & 0.967 & 0.790 & 0.755 &  1.013 & 0.063 & 0.986 & 0.185 &  0.648 \\
        \cellcolor{gray!15} & MASt3R-SLAM*~\cite{mast3rslam} & 0.070 & 0.032 & 0.055 & \second{0.056} & 0.035 & 0.118 & 0.041 & 0.116 & 0.020 & 0.060 \\
        \cellcolor{gray!15} & VGGT-SLAM~\cite{vggtslam}  & \best{\textbf{0.063}} & 0.031 & 0.048 & 0.152 & \best{\textbf{0.023}} & 0.133 & 0.038 & \second{0.039} & 0.020 & 0.061 \\
        \cellcolor{gray!15} & ViSTA-SLAM~\cite{vistaslam}   & 0.104 & \second{0.030} & \second{0.030} & 0.070 & 0.052 & \second{0.067} & \second{0.023} & 0.080 & \second{0.015} & \second{0.052} \\
        \multirow{-6}{*}{\cellcolor{gray!15}\rotatebox{90}{\textit{UnCalib.}}}  & UniSim-SLAM & \second{0.067} & \best{\textbf{0.018}} & \best{\textbf{0.022}} & \best{\textbf{0.034}} & \second{0.029} & \best{\textbf{0.056}} & \best{\textbf{0.021}} & \best{\textbf{0.031}} & \best{\textbf{0.013}} & \best{\textbf{0.032}} \\
        
        \bottomrule
    \end{tabular}
        }
\end{table*}

\begin{table*}[t]
    \centering
    \caption{\textbf{Quantitative trajectory results on 7-Scenes} (ATE RMSE [m]$\downarrow$)). 
    }
    \vspace{-2mm}
    \label{tab:ate_7scenes}
    \small
    \resizebox{1.0\linewidth}{!}{
    \setlength{\tabcolsep}{8pt}
        \begin{tabular}{c  l | *{7}{c} | c}
        \toprule
         & Method & chess & fire & heads & office & pumpkin & kitchen & stairs & Avg. \\
        \midrule
        
        \cellcolor{gray!15} & \textcolor{gray}{DROID-SLAM~\cite{droidslam}} & \textcolor{gray}{\textbf{0.018}} & \textcolor{gray}{\textbf{0.027}} & \textcolor{gray}{\textbf{0.021}} & \textcolor{gray}{\textbf{0.041}} & \textcolor{gray}{\textbf{0.025}} & \textcolor{gray}{\textbf{0.016}} & \textcolor{gray}{\textbf{0.017}} & \textcolor{gray}{\textbf{0.024}} \\
        \multirow{-2}{*}{\cellcolor{gray!15}\rotatebox{90}{\textit{Calib.}}} & \textcolor{gray}{MASt3R-SLAM~\cite{mast3rslam}} & \textcolor{gray}{0.082} & \textcolor{gray}{0.030} & \textcolor{gray}{0.024} & \textcolor{gray}{0.052} & \textcolor{gray}{0.050} & \textcolor{gray}{0.044} & \textcolor{gray}{0.027} & \textcolor{gray}{0.044} \\
        
        \midrule
        \midrule
        
        \cellcolor{gray!15} & CUT3R~\cite{cut3r} & 0.514 & 0.110 & 0.197 & 0.430 & 0.346 & 0.202 & 0.385 & 0.312 \\
        \cellcolor{gray!15} & SLAM3R~\cite{slam3r} & 0.131 & 0.044 & 0.040 & 0.058 & 0.100 & 0.064 & 0.116 & 0.079 \\
        \cellcolor{gray!15} & MASt3R-SLAM~\cite{mast3rslam} & 0.090 & 0.058 & 0.039 & 0.072 & 0.084 & 0.062 & 0.071 & 0.068 \\
        \cellcolor{gray!15} & VGGT-SLAM~\cite{vggtslam} & \second{0.039} & \second{0.024} & 0.041 & \second{0.032} & \second{0.050} & \second{0.034} & 0.042 & \second{0.037} \\
        \cellcolor{gray!15} & ViSTA-SLAM~\cite{vistaslam} & 0.075 & 0.035 & \second{0.030} & 0.064 & 0.065 & 0.041 & \second{0.036} & 0.049 \\
        \multirow{-6}{*}{\cellcolor{gray!15}\rotatebox{90}{\textit{Uncalib.}}} & UniSim-SLAM & \best{\textbf{0.017}} & \best{\textbf{0.018}} & \best{\textbf{0.026}} & \best{\textbf{0.024}} & \best{\textbf{0.022}} & \best{\textbf{0.016}} & \best{\textbf{0.019}} & \best{\textbf{0.020}} \\
        \bottomrule
        \end{tabular}
    }
\end{table*}
\subsection{Evaluation}
\begin{figure}[t]
    \centering
    \includegraphics[width=0.92\linewidth]{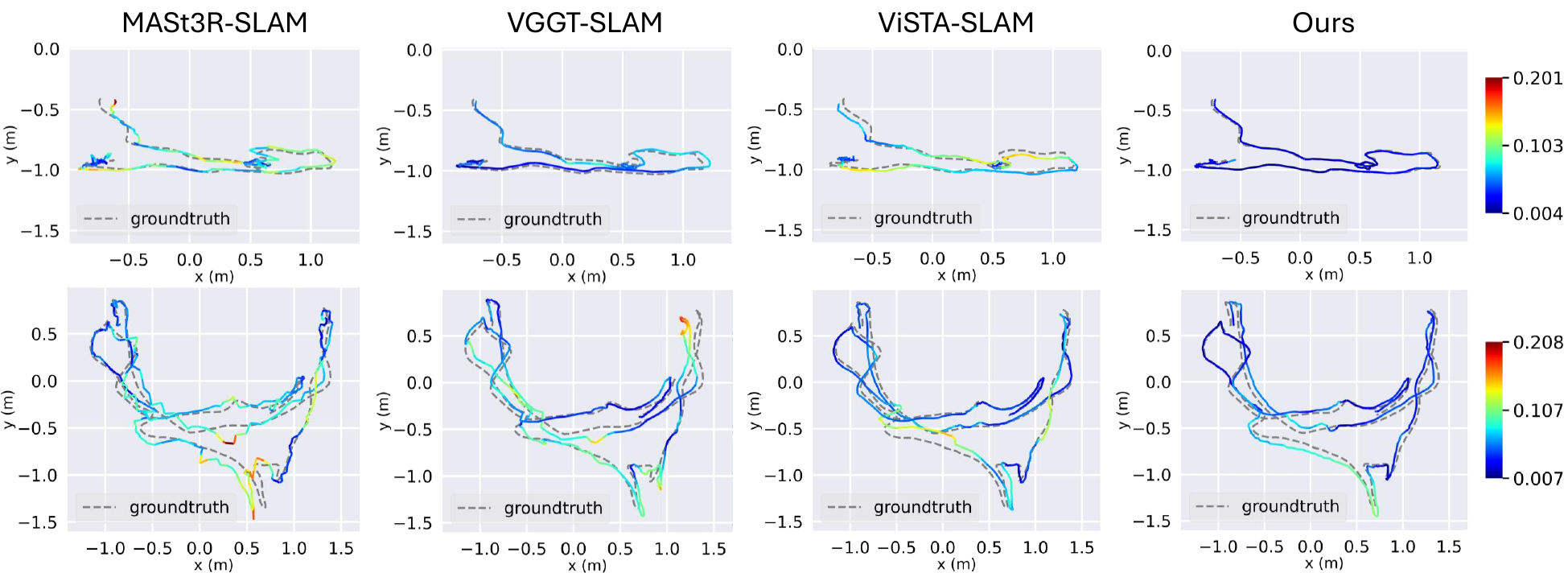}
    \caption{\textbf{Qualitative trajectory comparisons on 7-Scenes~\cite{7scenes}}}
    \label{fig:trajectory}
\end{figure}

\noindent \textbf{Camera Trajectory.} \ 
%
\texttt{UniSim-SLAM} achieves state-of-the-art performance in the uncalibrated setting on both TUM RGB-D (Tab.~\ref{tab:ate_tumrgbd}) and 7-Scenes (Tab.~\ref{tab:ate_7scenes}). 
On TUM RGB-D, we observe a noticeable reduction in trajectory error on the \textit{floor} sequence.
Because this scene is dominated by planar structures, it provides limited geometric cues for scale recovery, making feed-forward predictions prone to scale drift.
Our method significantly improves trajectory accuracy in this challenging case, indicating that the proposed multi-level factor graph effectively stabilizes scale by jointly optimizing view and submap constraints (see Fig.~\ref{fig:trajectory}).

A similar trend is observed on 7-Scenes.
In the \textit{chess} sequence, large depth variations and changes in camera-to-object distance introduce scale inconsistencies, highlighting the sensitivity of feed-forward models to input view configurations.
\texttt{UniSim-SLAM} substantially reduces trajectory error on this sequence, demonstrating that unified $Sim(3)$ optimization effectively mitigates scale drift.

\vspace{1mm} \noindent \textbf{3D Reconstruction.} \
To evaluate 3D reconstruction performance, we reconstruct a global point cloud using the optimized global view poses $\{T_i\}$ together with the multi-view depth estimates $\hat{D}^{\mathrm{mv}}$ and intrinsics, and confidence maps. 
Following \cite{vggtslam}, we discard points corresponding to the lowest 25\% confidence scores. 
Because our framework produces overlapping submaps, naively aggregating all reconstructed points would artificially improve completion metrics due to increased point density. 
To avoid this bias, we use, for each view, only the depth map from the submap with the highest confidence score.

\begin{figure}
    \centering
    \includegraphics[width=0.90\linewidth]{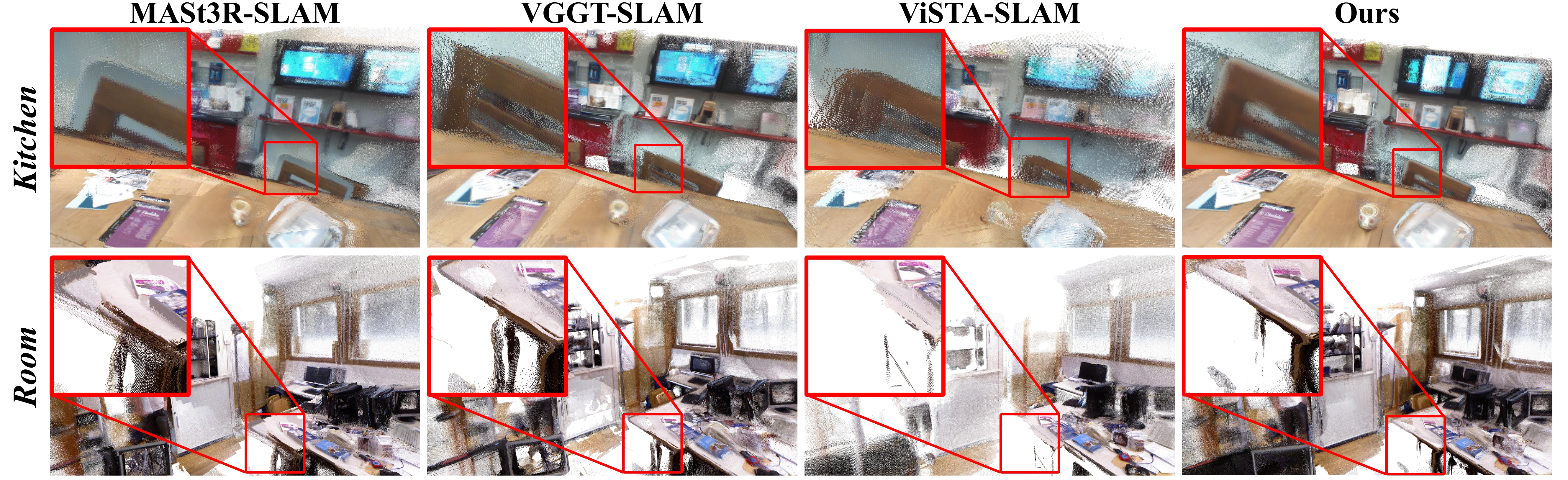}
    \caption{\textbf{Reconstruction results on 7-Scenes and TUM RGB-D.}  
    Red boxes indicate close-up views highlighting scene details.
    }
    \label{fig:reconstruction}
\end{figure}

Tab.~\ref{tab:recon_error} shows that \texttt{UniSim-SLAM} achieves lower Accuracy and Chamfer distance while maintaining comparable Completion, indicating improved geometric precision of the reconstructed scene. 
Fig.~\ref{fig:reconstruction} presents qualitative results on the 7-Scenes \textit{kitchen} and TUM RGB-D \textit{room} sequences.

\vspace{1mm} \noindent \textbf{Latency.} \ 
Tracking latency is an important aspect of SLAM systems, as camera pose estimates are expected to be available with low latency once input frames arrive. We therefore analyze the latency required to obtain a pose estimate during tracking. Specifically, we measure the elapsed time from receiving the minimum required input frames to producing a camera pose estimate in the frontend, and report the results in Tab.~\ref{tab:two_tables_ablation_latency_b}.
We observe that \texttt{UniSim-SLAM} achieves lower ATE by leveraging multi-view information while operating at a moderate latency. This suggests that the proposed system alleviates the accuracy–efficiency trade-off, although the latency remains higher than that of specialized two-view SLAM pipelines. This difference mainly arises from the large model size and general-purpose nature of our frontend backbone, VGGT.
To further analyze this aspect, we additionally evaluate a variant that replaces the frontend with a low-latency model (Ours + STA in Tab.~\ref{tab:two_tables_ablation_latency_b}). Although using different models in the frontend and backend introduces additional $Sim(3)$ inconsistencies, the resulting system still achieves better performance than existing methods. These results indicate that our multi-level factor graph can effectively perform $Sim(3)$ refinement even when integrating predictions from heterogeneous feed-forward models.
\begin{table}[t]
\centering
\caption{\textbf{Reconstruction error on 7-Scenes}. We report Accuracy (Acc.), Completeness (Comp.), and Chamfer distance. 
}
\label{tab:recon_error}
\scriptsize
\setlength{\tabcolsep}{4pt}
\renewcommand{\arraystretch}{1.2}
\resizebox{0.6\linewidth}{!}{
\begin{tabular}{c l|ccc}
    \toprule
    & Method & Acc. $\downarrow$ & Comp. $\downarrow$ & Chamfer $\downarrow$ \\
    \midrule
    \cellcolor{gray!15} & \textcolor{gray}{DROID-SLAM~\cite{droidslam}} & \textcolor{gray}{0.111} & \textcolor{gray}{\textbf{0.049}} & \textcolor{gray}{0.080} \\
    \multirow{-2}{*}{\cellcolor{gray!15}\rotatebox{90}{\textit{Calib.}}}& \textcolor{gray}{MASt3R-SLAM~\cite{mast3rslam}} & \textcolor{gray}{\textbf{0.064}} & \textcolor{gray}{0.068} & \textcolor{gray}{\textbf{0.066}} \\
    \midrule
    \midrule
    \cellcolor{gray!15} & Spann3R @5~\cite{spann3r} & 0.095 & \best{\textbf{0.041}} & 0.068 \\
    \cellcolor{gray!15} & CUT3R~\cite{cut3r} & 0.107 & 0.059 & 0.083 \\
    \cellcolor{gray!15} & SLAM3R~\cite{slam3r} & 0.069 & 0.153 & 0.111 \\
    \cellcolor{gray!15} & MASt3R-SLAM~\cite{mast3rslam} & 0.054 & 0.048 & 0.051 \\
    \cellcolor{gray!15} & VGGT-SLAM~\cite{vggtslam} & \second{0.039} & 0.051 & \second{0.045} \\
    \cellcolor{gray!15} & ViSTA-SLAM~\cite{vistaslam} & 0.041 & 0.056 & 0.049 \\
    \multirow{-8}{*}{\cellcolor{gray!15}\rotatebox{90}{\textit{Uncalib.}}} & UniSim-SLAM & \best{\textbf{0.035}} & \second{0.046} & \best{\textbf{0.041}} \\
\bottomrule
\end{tabular}
}
\end{table}

\subsection{Ablation study}
\noindent \textbf{Impact of Submap Configuration.} \ 
In Tab.~\ref{tab:two_tables_ablation_submap_a}, we analyze two key hyperparameters: the submap size $w$, which determines the number of keyframes per submap, and the submap overlap $\phi$, which controls the number of shared keyframes between adjacent submaps.

\noindent \textit{Submap Overlap.} \
As shown in Tab.~\ref{tab:two_tables_ablation_submap_a}, an overlap of a single frame ($\phi=1$) makes submap alignment rely entirely on a single multi-view local pose estimation, risking optimization instability. To prevent this, we adopt $\phi=2$ as the default configuration, ensuring robust optimization with minimal overhead.

\noindent \textit{Submap Size.} \
Larger submaps improve local multi-view estimation but reduce the number of submap nodes in the graph, weakening global constraints. 
Accordingly, large submaps ($w=32$) do not necessarily improve performance. 
We thus set $w=16$ by default. Additional experiment and details are in the supplementary material.

\vspace{1mm}
\noindent \textbf{Effectiveness of Graph Constraints.} \ 
In Tab~\ref{tab:ablation_graph_features}, we provide an ablation study evaluating the contribution of each backend component.
Under the default configuration ($\phi=2$), removing any individual constraint leads to a noticeable performance drop, indicating that the proposed multi-level factor graph effectively refines the global trajectory.
This confirms that the improvements arise from the synergistic interaction of multiple constraints rather than a single dominant factor.
Notably, even without loop closure, the system achieves performance comparable to existing state-of-the-art methods, suggesting that the proposed graph alone produces a globally consistent optimization.
Furthermore, in a non-overlap setting ($\phi=0$), the temporal view-to-view edge becomes particularly crucial, as it maintains graph connectivity and enables scale corrections to propagate across the trajectory.

\begin{table}[t]
\centering
\caption{\textbf{Ablation studies and latency comparison.}}
\vspace{-3mm}
\label{tab:two_tables_ablation_runtime}
\scriptsize

\begin{subtable}[t]{\linewidth}
    \centering
    \caption{Ablation Studies on 7-Scenes.}
    \vspace{-1mm}
    \label{tab:ablation_graph_features}
    \setlength{\tabcolsep}{4pt}
    \renewcommand{\arraystretch}{1.2}
    \begin{tabular}{c | cccccccc}
        \toprule
        & $w/o$ Backend
        & $w/o$ LC
        & $w/o\,\,\mathbf{e}^{\text{2v}}$
        & $w/o\,\,\mathbf{e}^{\mathrm{anch}}$
        & $w/o\,\,\mathbf{e}^{\mathrm{br}}$
        & $w/o\,\,\mathbf{e}^{\mathrm{tie}}$
        & $w/o\,\,\mathbf{e}^{\mathrm{sc}}$
        & Ours (full) \\
        \midrule
        $\phi=0$ & 0.124 & 0.061 & 0.101 & 0.083 & 0.116 & 0.040 & 0.048 & 0.032 \\
        $\phi=2$ & 0.124 & 0.037 & 0.021   & 0.020     & 0.063     & 0.027     &  0.031     & 0.020 \\
        \bottomrule
    \end{tabular}
\end{subtable}
 \par\smallskip
\begin{subtable}[t]{0.40\linewidth}
    \centering
    \caption{$w$ and $\phi$ on TUM RGB-D.}
    \vspace{-1mm}
    \label{tab:two_tables_ablation_submap_a}
    \setlength{\tabcolsep}{5.1pt}
    \renewcommand{\arraystretch}{1.2}
    \begin{tabular}{c | cccc}
        \toprule
        $w$ & $\phi=1$ & $\phi=2$ & $\phi=4$ & $\phi=8$ \\
        \midrule
        4  & 0.043 & 0.041 & -- & -- \\
        8  & 0.035 &   0.035   & 0.035 & -- \\
        16 & 0.031 & 0.032 & 0.031 & 0.030 \\
        32 & 0.033 & 0.032 & 0.034 & 0.033 \\
        \bottomrule
    \end{tabular}
\end{subtable}
\hspace{5mm}
\begin{subtable}[t]{0.45\linewidth}
    \centering
    \caption{Latency comparison on 7-Scenes.}
    \vspace{-1mm}
    \label{tab:two_tables_ablation_latency_b}
    \setlength{\tabcolsep}{2pt}
    \renewcommand{\arraystretch}{1.45}
    \resizebox{\linewidth}{!}{%
    \begin{tabular}{l | c|cc}
        \toprule
        Method & Frontend & Latency [ms]$\downarrow$ &  ATE$\downarrow$ \\
        \midrule
        MASt3R-SLAM~\cite{mast3rslam} & MASt3R~\cite{mast3r} & \second{90}  & 0.068 \\
        VGGT-SLAM~\cite{vggtslam}     & VGGT~\cite{vggt}     & 3410  & 0.037 \\
        ViSTA-SLAM~\cite{vistaslam}   & STA~\cite{vistaslam} & \best{\textbf{35}}  & 0.049 \\
        Ours   + STA~\cite{vistaslam}  & STA~\cite{vistaslam}     & \best{\textbf{35}}  & \second{0.027} \\
        Ours                          & VGGT~\cite{vggt}     & 197  & \best{\textbf{0.020}} \\
        \bottomrule
    \end{tabular}%
    }
\end{subtable}
\end{table}


%% file: sec/5_conclusion.tex
\section{Conclusion}

We proposed \texttt{UniSim-SLAM}, a feed-forward SLAM system that unifies low-latency two-view tracking and multi-view submaps within a $Sim(3)$ optimization framework. Our key insight is that predictions from different feed-forward inference regimes reside in heterogeneous coordinate systems and must therefore be jointly optimized.
By representing these predictions as complementary constraints in a multi-level factor graph, \texttt{UniSim-SLAM} enables consistent global pose estimation while preserving immediate tracking updates. This unified formulation bridges fast two-view inference and constraint-rich multi-view reconstruction, enabling robust drift correction.
Experiments on TUM RGB-D and 7-Scenes demonstrate that \texttt{UniSim-SLAM} outperforms existing feed-forward SLAM methods, highlighting the effectiveness of unified optimization for integrating heterogeneous geometric predictions.

%% file: supple.tex
\appendix 

\newcommand{\makesupptitle}[1]{%
  \clearpage
  \begin{center}%
    {\Large \bfseries\boldmath #1 \par}\vskip .8cm
    {\large Supplementary Material \par}\vskip .4cm
  \end{center}%
  \vskip .6cm
  
  \setcounter{section}{0}
  \setcounter{equation}{0}
  \setcounter{figure}{0}
  \setcounter{table}{0}
  \renewcommand{\thesection}{A\arabic{section}} 
  \renewcommand{\theequation}{S\arabic{equation}} 
  \renewcommand{\thefigure}{S\arabic{figure}} 
  \renewcommand{\thetable}{S\arabic{table}} 
}


\clearpage
\setcounter{page}{1}
\makesupptitle{UniSim-SLAM: Feed-Forward SLAM\\ with Unified $Sim(3)$ Optimization}





\definecolor{bestgreen}{RGB}{146,208,80}
\definecolor{secondorange}{RGB}{255,192,0}
\definecolor{calibgray}{gray}{0.5}

\definecolor{bestPink}{rgb}{0.980, 0.502, 0.447} 
\definecolor{secondOrange}{rgb}{1.000, 0.784, 0.486}
\colorlet{colorFst}{bestPink!40} 
\colorlet{colorSnd}{secondOrange!45} 
\colorlet{colorTrd}{Goldenrod!20} 

\renewcommand{\thefigure}{A\arabic{figure}}
\renewcommand{\thetable}{A\arabic{table}}
\renewcommand{\theequation}{A\arabic{equation}}
\renewcommand{\thesection}{A\arabic{section}}


\input{supple/0_overview}
\input{supple/1_Implementation_Details}

\input{supple/2_ablation_study}
\input{supple/3_addtional_experiments}

\input{supple/4_addtional_results}
\clearpage


%
%

%% file: supple/0_overview.tex
\section*{Overview}

The supplementary material provides additional details and experimental results that complement the main paper:
\begin{itemize}
    \item Sec.~\ref{sec:implementation_details} elaborates on the implementation details of $\texttt{UniSim-SLAM}$, including the backend optimization procedure for the unified $Sim(3)$ factor graph, the design of the factor weighting scheme, and the loop closure mechanism.

    \item Sec.~\ref{sec:ablation_study} presents an extended ablation study of the contributions of individual components in our framework, with a particular focus on the roles of different graph constraints and the robustness of the depth-based scale estimation.

    \item Sec.~\ref{sec:additional_experiments} reports additional experiments with alternative frontend backbones and a detailed runtime analysis.

    \item Sec.~\ref{sec:additional_results} provides additional quantitative results on Replica~\cite{replica}, together with qualitative results illustrating the effect of pose graph optimization and extended visualizations of reconstructed scenes on multiple datasets.
    




\end{itemize}

\appendix
\renewcommand{\thefigure}{A\arabic{figure}}
\renewcommand{\thetable}{A\arabic{table}}
\renewcommand{\theequation}{A\arabic{equation}}

%% file: supple/1_Implementation_Details.tex
\section{Implementation Details}
\label{sec:implementation_details}

\subsubsection{Factor graph weights.} \
For stable optimization of the unified factor graph, we use fixed residual weights for each constraint type across all experiments. The weights are defined as 
\begin{equation*}
(w^\text{2v}, w^\text{br}, w^\text{anch}, w^\text{sc}, w^\text{tie}) = (1.0, 1.0, 50.0, 50.0, 0.2).
\end{equation*}
Here $w^\text{2v}$ corresponds to the temporal view-to-view constraint and $w^\text{br}$ denotes the view-to-submap bridge constraint.
The terms $w^\text{anch}$ and $w^\text{sc}$ represent the scale anchoring and submap scale consistency constraints, respectively, while $w^\text{tie}$ controls the weight of the submap pose consistency constraint.
We assign larger weights to the scale-related constraints ($w^\text{anch}$, $w^\text{sc}$) to strongly enforce scale consistency during optimization, which helps stabilize the relative scale between the global trajectory and the submap coordinate frames.

In contrast, we assign a relatively small weight to the tie constraint. Rather than relying on direct submap-to-submap alignment, the optimization prioritizes view-to-submap constraints when integrating local submap predictions into the global trajectory.
%
The effect of this design is further examined in the ablation study. Tab.~\ref{tab:factor_graph_weight_supple} reports the performance when the scale-related weights are reduced or the tie constraint is assigned a larger weight. The results show that weakening the scale constraints or overemphasizing the tie constraint degrades performance.

\subsubsection{Loop Closure.} \
Loop candidates are retrieved using image descriptors extracted by SALAD~\cite{salad}.
For each query keyframe, we retrieve the most similar past keyframe based on descriptor similarity. The top-ranked retrieval is accepted as a loop candidate only if its similarity score exceeds a predefined threshold.
We then perform geometric verification to rule out a visually similar yet geometrically inconsistent match.
During this step, the query and retrieved keyframes are processed by a feed-forward model (VGGT~\cite{vggt}) to estimate their relative geometry.
The candidate is rejected if they exhibit implausibly large relative translation or insufficient field-of-view overlap between the two views.
Once a loop candidate passes this verification, we construct a joint loop submap around the matched keyframes.
Multi-view inference on this submap generates additional geometric constraints, which are then integrated into the unified $Sim(3)$ factor graph to refine the global trajectory.

\begin{table*}[t]
    \centering
    \caption{\textbf{Ablation study of factor graph weights on 7-Scenes} (ATE RMSE [m]$\downarrow$). 
    }
    \vspace{-2mm}
    \label{tab:factor_graph_weight_supple}
    \small
    \resizebox{1.0\linewidth}{!}{
    \setlength{\tabcolsep}{8pt}
        \begin{tabular}{ c | c | *{7}{c} | c}
        \toprule
        Residual & weights & chess & fire & heads & office & pumpkin & kitchen & stairs & Avg. \\
        \midrule

        \multirow{4}{*}{$\mathbf{e}^\text{sc},\mathbf{e}^\text{anch}$}
        & $w^\text{sc}, w^\text{anch} = 1$ & 0.030 & 0.024 & 0.027 & 0.027 & 0.021 & 0.022 & 0.028 & 0.025 \\
        & $w^\text{sc}, w^\text{anch} = 2$ & 0.024 & 0.022 & 0.027 & 0.026 & 0.021 & 0.021 & 0.025 & 0.024 \\
        & $w^\text{sc}, w^\text{anch} = 10$ & 0.018 & 0.019 & 0.027 & 0.025 & 0.021 & 0.019 & 0.020 & 0.021 \\
        & $w^\text{sc}, w^\text{anch} = 50$ & 0.017 & 0.018 & 0.026 & 0.024 & 0.022 & 0.016 & 0.019 & 0.020 \\
        
        \midrule
        
        \multirow{4}{*}{$\mathbf{e}^\text{tie}$}
        & $w^\text{tie} = 0.2$ & 0.017 & 0.018 & 0.026 & 0.024 & 0.022 & 0.016 & 0.019 & 0.020 \\
        & $w^\text{tie} = 0.5$ & 0.018 & 0.018 & 0.026 & 0.024 & 0.021 & 0.015 & 0.019 & 0.020 \\
        & $w^\text{tie} = 1.0$ & 0.020 & 0.020 & 0.026 & 0.023 & 0.020 & 0.015 & 0.019 & 0.020 \\
        & $w^\text{tie} = 2.0$ & 0.025 & 0.023 & 0.027 & 0.024 & 0.020 & 0.015 & 0.019 & 0.022 \\
        \bottomrule
        \end{tabular}
    }
\end{table*}

%% file: supple/2_ablation_study.tex
\section{Ablation Study}
\label{sec:ablation_study}
\subsubsection{Effectiveness of Graph Constraints.} \
Tab.~\ref{tab:with_without} extends the ablation study presented in Tab.~4a of the main paper and analyzes the contribution of different edge types in the proposed multi-level factor graph. In particular, we examine the effects of removing edges at different levels of the graph, including the submap-to-submap and view-to-submap connections.

Removing either type of edge degrades overall performance, indicating that the different levels of the factor graph interact in a complementary manner. Rather than relying on a single dominant constraint, the proposed graph structure benefits from the interaction among multiple edge types.

In particular, removing the view-to-submap edges disconnects the submaps from the global trajectory. As a result, the global pose nodes are no longer properly constrained by the submap predictions, and the system behaves similarly to a tracking-only system without backend refinement.

Furthermore, removing the submap-to-submap edges also causes a noticeable drop in performance. This indicates that direct constraints between neighboring submaps contribute to more stable pose refinement by enforcing consistency between independently estimated submap coordinate frames.

\begin{table*}[t]
    \centering
    \caption{\textbf{Ablation study of graph constraints on 7-Scenes} (ATE RMSE [m]$\downarrow$).}
    \vspace{-2mm}
    \label{tab:with_without}
    \small
    \resizebox{1.0\linewidth}{!}{
    \setlength{\tabcolsep}{8pt}
        \begin{tabular}{ c | l | *{7}{c} | c}
        \toprule
         & Method & chess & fire & heads & office & pumpkin & kitchen & stairs & Avg. \\
        \midrule

        \multirow{10}{*}{\rotatebox{90}{overlap ($\phi=2$)}}
        & Tracking only   & 0.158 & 0.090 & 0.101 & 0.145 & 0.166 & 0.114 & 0.096 & 0.124 \\
        & w/o Loop closure & 0.020 & 0.024 & 0.029 & 0.075 & 0.036 & 0.050 & 0.023 & 0.037 \\
        & w/o$\,\,\mathbf{e}^\text{sc}$ & 0.045 & 0.032 & 0.027 & 0.030 & 0.023 & 0.022 & 0.036 & 0.031 \\
        & w/o$\,\,\mathbf{e}^\text{2v}$ & 0.020 & 0.018 & 0.026 & 0.024 & 0.023 & 0.016 & 0.019 & 0.021 \\
        & w/o$\,\,\mathbf{e}^\text{anch}$ & 0.019 & 0.018 & 0.026 & 0.024 & 0.022 & 0.015 & 0.018 & 0.020 \\
        & w/o$\,\,\mathbf{e}^\text{br}$ & 0.046 & 0.048 & 0.035 & 0.110 & 0.121 & 0.051 & 0.027 & 0.063 \\
        & w/o$\,\,\mathbf{e}^\text{tie}$ & 0.018 & 0.020 & 0.026 & 0.035 & 0.033 & 0.034 & 0.021 & 0.027 \\
        & w/o$\,\,\mathcal{E}^\text{v2s}$ & 0.158 & 0.090 & 0.101 & 0.145 & 0.166 & 0.114 & 0.096 & 0.124 \\
        & w/o$\,\,\mathcal{E}^\text{s2s}$ & 0.017 & 0.023 & 0.027 & 0.037 & 0.030 & 0.052 & 0.025 & 0.030 \\
        & Ours (full) & 0.017 & 0.018 & 0.026 & 0.024 & 0.022 & 0.016 & 0.019 & 0.020 \\
        
        \midrule

        \multirow{10}{*}{\rotatebox{90}{Non-overlap}}
        & Tracking only   & 0.158 & 0.090 & 0.101 & 0.145 & 0.166 & 0.114 & 0.096 & 0.124 \\
        & w/o Loop closure & 0.106 & 0.045 & 0.035 & 0.074 & 0.075 & 0.059 & 0.033 & 0.061 \\
        & w/o$\,\,\mathbf{e}^\text{sc}$ & 0.117 & 0.056 & 0.032 & 0.033 & 0.037 & 0.033 & 0.030 & 0.048 \\
        & w/o$\,\,\mathbf{e}^\text{2v}$ & 0.157 & 0.064 & 0.059 & 0.065 & 0.162 & 0.111 & 0.085 & 0.101 \\
        & w/o$\,\,\mathbf{e}^\text{anch}$ & 0.178 & 0.185 & 0.070 & 0.047 & 0.020 & 0.042 & 0.040 & 0.083 \\
        & w/o$\,\,\mathbf{e}^\text{br}$ & 0.216 & 0.098 & 0.075 & 0.138 & 0.125 & 0.112 & 0.045 & 0.116 \\
        & w/o$\,\,\mathbf{e}^\text{tie}$ & 0.106 & 0.043 & 0.041 & 0.024 & 0.020 & 0.033 & 0.015 & 0.040 \\
        & w/o$\,\, \mathcal{E}^\text{v2s}$ & 0.158 & 0.090 & 0.101 & 0.145 & 0.166 & 0.114 & 0.096 & 0.124 \\
        & w/o$\,\, \mathcal{E}^\text{s2s}$ & 0.092 & 0.029 & 0.033 & 0.027 & 0.034 & 0.045 & 0.025 & 0.041 \\
        & Ours (full) & 0.066 & 0.029 & 0.043 & 0.027 & 0.023 & 0.019 & 0.015 & 0.032 \\
        
        \bottomrule
        \end{tabular}
    }
\end{table*}

\begin{table*}[t]
    \centering
    \caption{\textbf{Ablation study of scale estimation on 7-Scenes} (ATE RMSE [m]$\downarrow$).}
    \vspace{-2mm}
    \label{tab:depthmap}
    \small
    \resizebox{0.4\linewidth}{!}{
    \setlength{\tabcolsep}{8pt}
        \begin{tabular}{ l | c | c}
        \toprule
         Target & depth noise ($\sigma$) & Avg. \\
        \midrule

        \multirow{4}{*}{\emph{Scale}}
        & $\sigma = 0.15$ & 0.045 \\
        & $\sigma = 0.1$ & 0.041 \\
        & $\sigma = 0.05$ & 0.026 \\
        & $\sigma = 0$ & 0.020 \\

        \midrule

        \multirow{4}{*}{\emph{Depth}}
        & $\sigma = 0.3$ & 0.021 \\
        & $\sigma = 0.2$ & 0.021 \\
        & $\sigma = 0.1$ & 0.020 \\
        & $\sigma = 0$ & 0.020 \\

        \bottomrule
        \end{tabular}
    }
\end{table*}

\subsubsection{Robustness of Depth-based Scale Estimation.} \ 
In Tab.~\ref{tab:depthmap}, we evaluate the robustness of the depth-based relative scale estimation used in $\texttt{UniSim-SLAM}$.
We analyze the effect of inaccurate scale estimates by perturbing the estimated scale with synthetic noise (denoted as \emph{Scale} in the table). Specifically, we inject Gaussian noise into the depth-derived relative scale, where the noise magnitude is controlled by the standard deviation $\sigma$. As the noise level increases, the performance degrades consistently, confirming that inaccurate scale estimates negatively affect overall performance.

To further analyze this behavior, we conduct an additional ablation study by corrupting the depth map itself (denoted as \emph{Depth} in the table). Specifically, we add Gaussian noise to the depth values and randomly remove $10\%$ of the pixels to simulate inaccurate geometric predictions from the feed-forward model. Although scale estimation in $\texttt{UniSim-SLAM}$ depends on depth quality, the scale is computed from pixel-aligned correspondences since the depth maps originate from the same view. As a result, the system remains robust even when $\sigma$ ranges from 0.1 to 0.3.

%% file: supple/3_addtional_experiments.tex
\section{Additional Experiments}
\label{sec:additional_experiments}

\begin{table*}[t]
    \centering
    \caption{\textbf{Additional trajectory results on TUM RGB-D} (ATE RMSE $[m]$ $\downarrow$). ‘$\times$’ denotes failure to produce a valid trajectory and the * symbol denotes uncalibrated methods.}
    \label{tab:additional_ate_tumrgbd}
    \vspace{-2mm}
    \small
    \resizebox{1.0\linewidth}{!}{
    \setlength{\tabcolsep}{8pt} 

    \begin{tabular}{ll|ccccccccc|c}
        \toprule
        & Method & 360 & desk & desk2 & floor & plant & room & rpy & teddy & xyz & Avg \\
        \midrule
        
        \cellcolor{gray!15} & \calib{ORB-SLAM3~\cite{orbslam3}}    & \calib{$\times$} & \calib{0.017} & \calib{0.210} & \calib{$\times$} & \calib{0.034} & \calib{$\times$} & \calib{$\times$} & \calib{$\times$} & \calib{\textbf{0.009}} & \calib{N/A} \\
        \cellcolor{gray!15} & \calib{DeepV2D~\cite{deepv2d}}     & \calib{0.243} & \calib{0.166} & \calib{0.379} & \calib{1.653} & \calib{0.203} & \calib{0.246} & \calib{0.105} & \calib{0.316} & \calib{0.064} & \calib{0.375} \\
        \cellcolor{gray!15} & \calib{DeepFactors~\cite{deepfactors}} & \calib{0.159} & \calib{0.170} & \calib{0.253} & \calib{0.169} & \calib{0.305} & \calib{0.364} & \calib{0.043} & \calib{0.601} & \calib{0.035} & \calib{0.233} \\
        \cellcolor{gray!15} & \calib{DPV-SLAM~\cite{dpvslam}}    & \calib{0.112} & \calib{0.018} & \calib{0.029} & \calib{0.057} & \calib{0.021} & \calib{0.330} & \calib{0.030} & \calib{0.084} & \calib{0.010} & \calib{0.076} \\
        \cellcolor{gray!15} & \calib{DPV-SLAM++~\cite{dpvslam}}  & \calib{0.132} & \calib{0.018} & \calib{0.029} & \calib{0.050} & \calib{0.022} & \calib{0.096} & \calib{0.032} & \calib{0.098} & \calib{0.010} & \calib{0.054} \\
        \cellcolor{gray!15}& \calib{GO-SLAM~\cite{goslam}}     & \calib{0.089} & \calib{\textbf{0.016}} & \calib{0.028} & \calib{0.025} & \calib{0.026} & \calib{0.052} & \calib{\textbf{0.019}} & \calib{0.048} & \calib{0.010} & \calib{0.035} \\
        \cellcolor{gray!15} & \calib{DROID-SLAM~\cite{droidslam}}  & \calib{0.111} & \calib{0.018} & \calib{0.042} & \calib{\textbf{0.021}} & \calib{\textbf{0.016}} & \calib{\textbf{0.049}} & \calib{0.026} & \calib{0.048} & \calib{0.012} & \calib{0.038} \\
        \multirow{-8}{*}{\cellcolor{gray!15}\rotatebox{90}{\textit{Calib.}}} & \calib{MASt3R-SLAM~\cite{mast3rslam}} & \calib{\textbf{0.049}} & \calib{\textbf{0.016}} & \calib{\textbf{0.024}} & \calib{0.025} & \calib{0.020} & \calib{0.061} & \calib{0.027} & \calib{\textbf{0.041}} & \calib{\textbf{0.009}} & \calib{\textbf{0.030}} \\
        
        \midrule 
        \midrule
        
        \cellcolor{gray!15} & CUT3R~\cite{cut3r} & 0.174 & 0.592 & 0.546 & 0.662 & 0.467 & 0.911 & 0.051 & 0.845  & 0.129  & 0.486 \\
        \cellcolor{gray!15} & SLAM3R~\cite{slam3r} & 0.211 & 0.861 & 0.967 & 0.790 & 0.755 &  1.013 & 0.063 & 0.986 & 0.185 &  0.648 \\
        \cellcolor{gray!15} & MASt3R-SLAM*~\cite{mast3rslam} & 0.070 & 0.032 & 0.055 & 0.056 & 0.035 & 0.118 & 0.041 & 0.116 & 0.020 & 0.060 \\

        \cellcolor{gray!15} & VGGT-SLAM~\cite{vggtslam}  & \best{\textbf{0.063}} & 0.031 & 0.048 & 0.152 & \best{\textbf{0.023}} & 0.133 & 0.038 & 0.039 & 0.020 & 0.061 \\
        \cellcolor{gray!15} & ViSTA-SLAM~\cite{vistaslam}   & 0.104 & 0.030 & \second{0.030} & 0.070 & 0.052 & 0.067 & 0.023 & 0.080 & 0.015 & \second{0.052} \\
        \cellcolor{gray!15} & Ours + STA &
        \second{0.065} & \best{\textbf{0.016}} & \best{\textbf{0.022}} & \second{0.035} & 0.045 & \best{\textbf{0.041}} & \second{0.022} & \second{0.034} & \best{\textbf{0.012}} & \best{\textbf{0.032}} \\
        
        \multirow{-7}{*}{\cellcolor{gray!15}\rotatebox{90}{\textit{Uncalib.}}} & Ours & 0.067 & \second{0.018} & \best{\textbf{0.022}} & \best{\textbf{0.034}} & \second{0.029} & \second{0.056} & \best{\textbf{0.021}} & \best{\textbf{0.031}} & \second{0.013} & \best{\textbf{0.032}} \\
        
        \bottomrule
    \end{tabular}
        }
\end{table*}

\begin{table*}[t]
    \centering
    \caption{\textbf{Additional trajectory results on 7-Scenes} (ATE RMSE [m]$\downarrow$). 
    }
    \vspace{-2mm}
    \label{tab:additional_ate_7scenes}
    \small
    \resizebox{1.0\linewidth}{!}{
    \setlength{\tabcolsep}{8pt}
        \begin{tabular}{c  l | *{7}{c} | c}
        \toprule
         & Method & chess & fire & heads & office & pumpkin & kitchen & stairs & Avg. \\
        \midrule
        
        \cellcolor{gray!15} & \textcolor{gray}{DROID-SLAM~\cite{droidslam}} & \textcolor{gray}{\textbf{0.018}} & \textcolor{gray}{\textbf{0.027}} & \textcolor{gray}{\textbf{0.021}} & \textcolor{gray}{\textbf{0.041}} & \textcolor{gray}{\textbf{0.025}} & \textcolor{gray}{\textbf{0.016}} & \textcolor{gray}{\textbf{0.017}} & \textcolor{gray}{\textbf{0.024}} \\

        \multirow{-2}{*}{\cellcolor{gray!15}\rotatebox{90}{\textit{Calib.}}} & \textcolor{gray}{MASt3R-SLAM~\cite{mast3rslam}} & \textcolor{gray}{0.082} & \textcolor{gray}{0.030} & \textcolor{gray}{0.024} & \textcolor{gray}{0.052} & \textcolor{gray}{0.050} & \textcolor{gray}{0.044} & \textcolor{gray}{0.027} & \textcolor{gray}{0.044} \\
        
        \midrule
        \midrule
        
        \cellcolor{gray!15} & CUT3R~\cite{cut3r} & 0.514 & 0.110 & 0.197 & 0.430 & 0.346 & 0.202 & 0.385 & 0.312 \\
        \cellcolor{gray!15} & SLAM3R~\cite{slam3r} & 0.131 & 0.044 & 0.040 & 0.058 & 0.100 & 0.064 & 0.116 & 0.079 \\
        \cellcolor{gray!15} & MASt3R-SLAM*~\cite{mast3rslam} & 0.090 & 0.058 & 0.039 & 0.072 & 0.084 & 0.062 & 0.071 & 0.068 \\
        \cellcolor{gray!15} & VGGT-SLAM~\cite{vggtslam} & 0.039 & \second{0.024} & 0.041 & 0.032 & 0.050 & 0.034 & 0.042 & 0.037 \\
        \cellcolor{gray!15} & ViSTA-SLAM~\cite{vistaslam} & 0.075 & 0.035 & \second{0.030} & 0.064 & 0.065 & 0.041 & 0.036 & 0.049 \\
        \cellcolor{gray!15} & Ours + STA & \second{0.030} & 0.031 & 0.031 & \best{\textbf{0.023}} & \second{0.029} & \second{0.024} & \best{\textbf{0.018}} & \second{0.027} \\

        \multirow{-7}{*}{\cellcolor{gray!15}\rotatebox{90}{\textit{Uncalib.}}} & Ours & \best{\textbf{0.017}} & \best{\textbf{0.018}} & \best{\textbf{0.026}} & \second{0.024} & \best{\textbf{0.022}} & \best{\textbf{0.016}} & \second{0.019} & \best{\textbf{0.020}} \\
        \bottomrule
        \end{tabular}
    }
\end{table*}

\subsubsection{Frontend Backbone.} \
$\texttt{UniSim-SLAM}$ separates the frontend and backend into independent threads that operate asynchronously. As a result, the feed-forward models used in the two modules need not be identical. In particular, the frontend two-view model $f^{\text{2v}}$ and the backend multi-view model $f^{\text{mv}}$ can differ, since the global trajectory is refined through depth- and pose-based $Sim(3)$ optimization.

To evaluate this flexibility, we test a heterogeneous configuration in which the frontend uses a low-latency two-view feed-forward model STA~\cite{vistaslam}, while the backend retains the multi-view VGGT model~\cite{vggt}. Tabs.~\ref{tab:additional_ate_tumrgbd} and \ref{tab:additional_ate_7scenes} report the results of this configuration, denoted as Ours + STA. Although this setting inherits some limitations of STA-based SLAM (ViSTA-SLAM), especially on challenging sequences such as \textit{chess} in 7-Scenes and \textit{plant} in TUM RGB-D, the overall performance remains competitive. These results indicate that $\texttt{UniSim-SLAM}$ can operate with heterogeneous feed-forward models, allowing the frontend to be replaced with a lower-latency alternative.

\subsubsection{Runtime Analysis.} \ 
Tab.~\ref{tab:runtime_analysis} reports the latency of each stage in $\texttt{UniSim-SLAM}$.
The frontend performs two-view estimation for every incoming keyframe, achieving approximately 25 FPS on 7-Scenes.
Meanwhile, the backend is triggered only after a sufficient number of new keyframes have been accumulated.
Under the default configuration, backend processing begins once 14 new keyframes have been collected.
Because the backend runs in a separate thread, frontend tracking and backend refinement are processed asynchronously without blocking each other, enabling low-latency tracking performance.
Please refer to the supplementary video for a demonstration.
\begin{table*}[t]
    \centering
    \caption{\textbf{Latency of each stage in UniSim-SLAM}. }
    \label{tab:runtime_analysis}
    \vspace{-2mm}
    \resizebox{1.0\linewidth}{!}{
    \setlength{\tabcolsep}{5pt}
        \begin{tabular}{ c | c | c  c  c  c  c}
        \toprule
        & Frontend & \multicolumn{5}{c}{Backend} \\ 
        \midrule
        Component & \makecell{Two-view \\ inference} & \makecell{Submap node \\ initialization} & \makecell{Descriptor \\ extraction} & \makecell{Graph \\ construction} & Optimization & \makecell{Multi-view \\ inference} \\
        
        \midrule

        \rule[-5pt]{0pt}{15pt}
        Latency[ms] & 197 & 167 & 16 & 36 & 645 & 933 \\
        \bottomrule
        \end{tabular}
    }
\end{table*}

%% file: supple/4_addtional_results.tex
\section{Additional Results}
\label{sec:additional_results}

\subsubsection{Quantitative results on Replica.}
\begin{table*}[t]
    \centering
    \caption{\textbf{Trajectory error comparison on Replica} (ATE RMSE [m]$\downarrow$). 
    }
    \label{tab:ate_replica}
    \small   
    \vspace{-2mm}
    \resizebox{1.0\linewidth}{!}{
    \setlength{\tabcolsep}{8pt}
    \begin{tabular}{ c | cccccccc | c}
        \toprule
        Method & office0 & office1 & office2 & office3 & office4 & room0 & room1 & room2 & Avg. \\
        \midrule

        MASt3R-SLAM$^{*}$\cite{mast3rslam} & \best{\textbf{0.056}} & 0.056 & 0.079 & 0.056 & 0.059 & 0.102 & 0.108 & 0.063 & 0.072 \\
        VGGT-SLAM\cite{vggtslam} & \best{\textbf{0.056}} & \second{0.034} & \second{0.043} & 0.057 & \second{0.038} & \second{0.034} & \second{0.037} & \second{0.036} & \second{0.042} \\
        ViSTA-SLAM\cite{vistaslam} & 0.074 & 0.193 & 0.118 & \second{0.048} & 0.130 & 0.069 & 0.093 & 0.136 & 0.108 \\
        Ours & \second{0.068} & \best{\textbf{0.018}} & \best{\textbf{0.022}} & \best\textbf{{0.020}} & \best\textbf{{0.028}} & \best{\textbf{0.023}} & \best{\textbf{0.021}} & \best{\textbf{0.028}} & \best{\textbf{0.029}} \\

        \bottomrule
    \end{tabular}
    }
\end{table*}
Tab.~\ref{tab:ate_replica} reports camera trajectory evaluation on Replica\cite{replica}, using the rendered RGB-D sequences provided by iMAP\cite{imap}. 
To eliminate the effect of keyframe selection, we use a keyframe selection stride of 5 for all methods.
In addition, Tab.~\ref{tab:recon_replica_comparison} presents reconstruction results on Replica.

\begin{table*}[t]
    \centering
    \caption{\textbf{Reconstruction error on Replica}. We report Accuracy (Acc.), Completeness (Comp.), and Chamfer distance (RMSE [m]$\downarrow$).
    }
    \label{tab:recon_replica_comparison} 
    \vspace{-2mm}
    \resizebox{0.55\linewidth}{!}{
    \setlength{\tabcolsep}{8pt}
    \begin{tabular}{ c | c  c  c}
        \toprule
        method & Acc.$\downarrow$ & Comp.$\downarrow$ & Chamfer$\downarrow$ \\
        \midrule


        MASt3R-SLAM$^{*}$\cite{mast3rslam} & 0.118 & \second{0.027} & 0.072 \\
        VGGT-SLAM\cite{vggtslam} & \second{0.064} & 0.044 & \second{0.054} \\
        

        ViSTA-SLAM\cite{vistaslam} & 0.114 & 0.053 & 0.084 \\
        Ours & \best{\textbf{0.050}} & \best{\textbf{0.022}} & \best{\textbf{0.036}}  \\
        
        \bottomrule
    \end{tabular}
    }
\end{table*}
\begin{figure}
    \centering
    \includegraphics[width=0.90\linewidth]{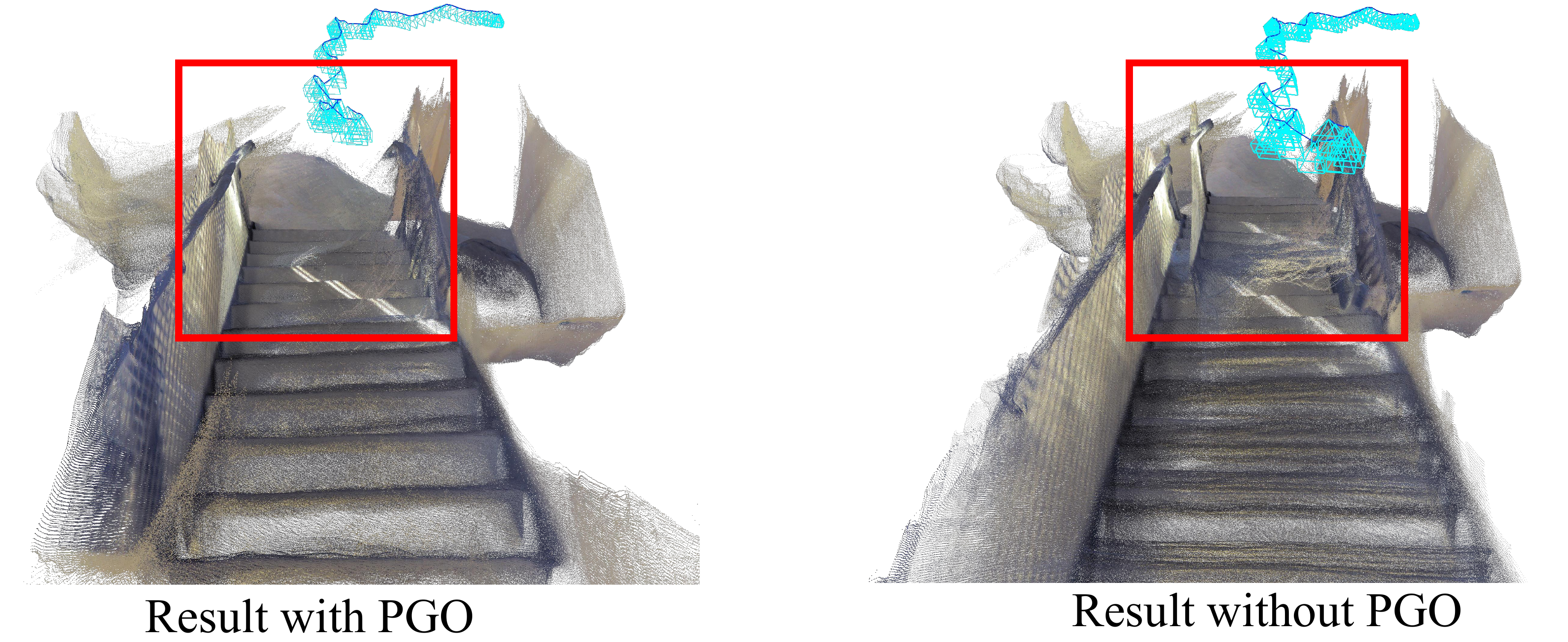}
    \vspace{-2mm}
    \caption{\textbf{Qualitative comparison with and without Pose Graph Optimization}. Red boxes highlight misaligned areas corrected by pose graph optimization. }
    \label{fig:Reconstruction_PGO_supple}
\end{figure}
\subsubsection{Pose Graph Optimization.} \
Fig.~\ref{fig:Reconstruction_PGO_supple} presents reconstruction and trajectory estimation results with and without pose graph optimization on the 7-Scenes \textit{stairs} sequence. The figure shows that pose graph optimization resolves misalignments across submaps and views.

\subsubsection{Additional Reconstruction Results.} 
\begin{figure}
    \centering
    \includegraphics[width=0.95\linewidth]{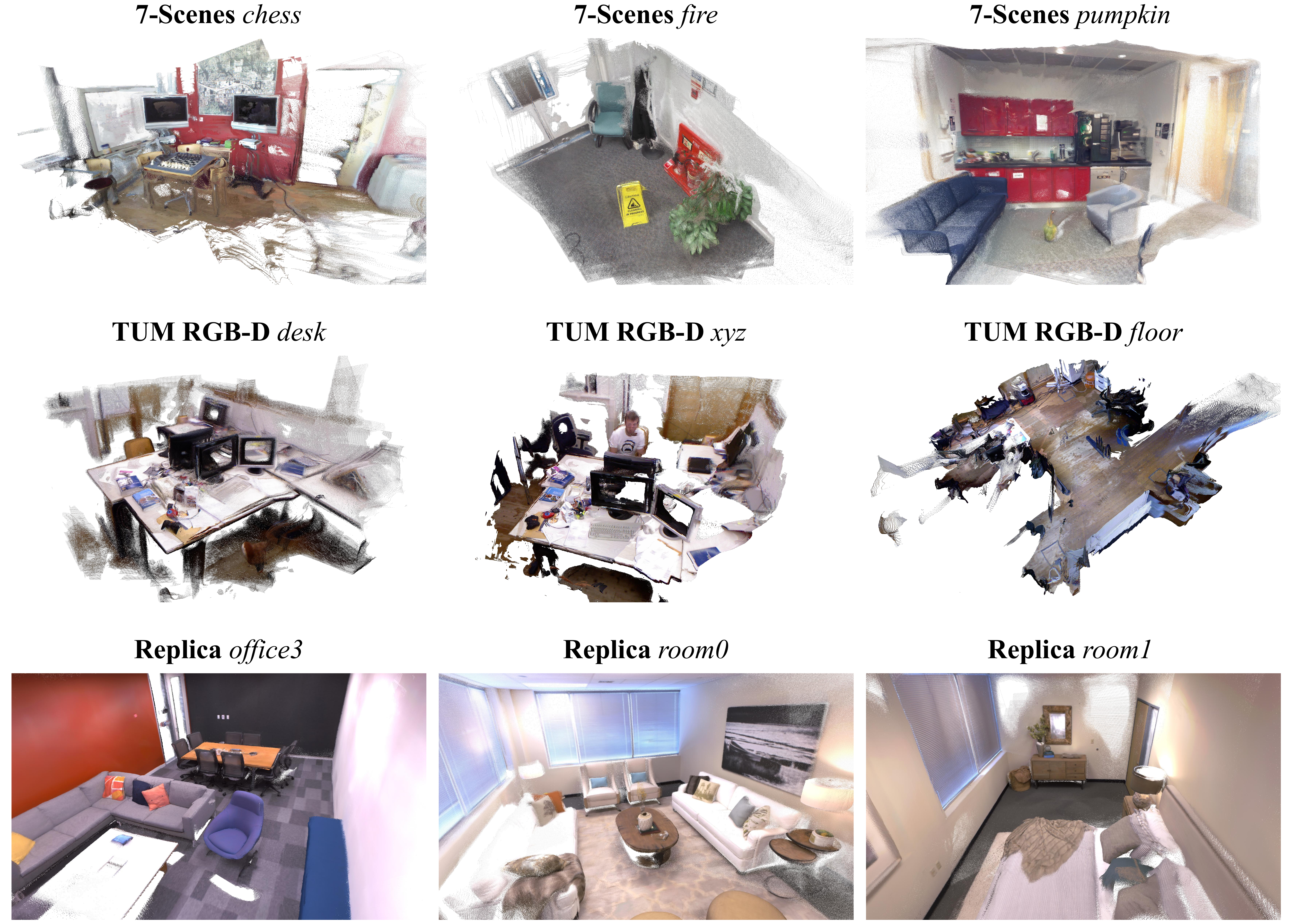}
    \vspace{-2mm}
    \caption{\textbf{Additional qualitative results} }
    \label{fig:Reconstruction_supple}
\end{figure}

\
Fig.~\ref{fig:Reconstruction_supple} presents additional reconstruction results on 7-Scenes, TUM RGB-D, and Replica. These results illustrate that $\texttt{UniSim-SLAM}$ performs stably across various scenes.

%% file: main.bib
@String(IJCV  = {Int. J. Comput. Vis.})

@String(CVPR  = {IEEE Conf. Comput. Vis. Pattern Recog.})

@String(ICCV  = {Int. Conf. Comput. Vis.})

@String(ECCV  = {Eur. Conf. Comput. Vis.})

@String(NeurIPS = {Adv. Neural Inform. Process. Syst.})

@String(ICLR  = {Int. Conf. Learn. Represent.})

@String(TMLR  = {Trans. Mach. Learn Res.})

@String(TRO    = {IEEE Trans. Robot.})

@String(RAL    = {IEEE Robot. Autom. Lett.})

@String(ICRA   = {IEEE Int. Conf. Robot. Autom.})

@String(IROS   = {IEEE/RSJ Int. Conf. Intell. Robots Syst.})

@String(TPAMI  = {IEEE Trans. Pattern Anal. Mach. Intell.})

@String(THREEDV= {Int. Conf. 3D Vis.})

@inproceedings{tumrgbd,
  author={Sturm, Jürgen and Engelhard, Nikolas and Endres, Felix and Burgard, Wolfram and Cremers, Daniel},
  booktitle=IROS, 
  title={A benchmark for the evaluation of RGB-D SLAM systems}, 
  year={2012},
  pages={573-580},
  doi={10.1109/IROS.2012.6385773}
}

@inproceedings{7scenes,
  author={Shotton, Jamie and Glocker, Ben and Zach, Christopher and Izadi, Shahram and Criminisi, Antonio and Fitzgibbon, Andrew},
  booktitle=CVPR, 
  title={Scene Coordinate Regression Forests for Camera Relocalization in RGB-D Images}, 
  year={2013},
  pages={2930-2937},
  doi={10.1109/CVPR.2013.377}
}

@article{sfm,
  title={Building rome in a day},
  author={Agarwal, Sameer and Furukawa, Yasutaka and Snavely, Noah and Simon, Ian and Curless, Brian and Seitz, Steven M and Szeliski, Richard},
  journal={Communications of the ACM},
  volume={54},
  number={10},
  pages={105--112},
  year={2011},
  publisher={ACM New York, NY, USA}
}

@inproceedings{incrementalsfm,
  title={Robust incremental structure-from-motion with hybrid features},
  author={Liu, Shaohui and Gao, Yidan and Zhang, Tianyi and Pautrat, R{\'e}mi and Sch{\"o}nberger, Johannes L and Larsson, Viktor and Pollefeys, Marc},
  booktitle=ECCV,
  pages={249--269},
  year={2024},
  organization={Springer}
}

@inproceedings{sfmrevisited,
  author={Schönberger, Johannes L. and Frahm, Jan-Michael},
  booktitle=CVPR, 
  title={Structure-from-Motion Revisited}, 
  year={2016},
  pages={4104-4113},
  doi={10.1109/CVPR.2016.445}
}

@inproceedings{pnp,
  author={Kneip, Laurent and Scaramuzza, Davide and Siegwart, Roland},
  booktitle=CVPR, 
  title={A novel parametrization of the perspective-three-point problem for a direct computation of absolute camera position and orientation}, 
  year={2011},
  pages={2969-2976},
  doi={10.1109/CVPR.2011.5995464}
}

@article{epnp,
  title={EP n P: An accurate O (n) solution to the P n P problem},
  author={Lepetit, Vincent and Moreno-Noguer, Francesc and Fua, Pascal},
  journal=IJCV,
  volume={81},
  number={2},
  pages={155--166},
  year={2009},
  publisher={Springer}
}

@article{orbslam,
  author={Mur-Artal, Raúl and Montiel, J. M. M. and Tardós, Juan D.},
  journal=TRO, 
  title={ORB-SLAM: A Versatile and Accurate Monocular SLAM System}, 
  year={2015},
  volume={31},
  number={5},
  pages={1147-1163},
  doi={10.1109/TRO.2015.2463671}
}

@article{orbslam2,
  author={Mur-Artal, Raúl and Tardós, Juan D.},
  journal=TRO, 
  title={{ORB-SLAM2}: An Open-Source SLAM System for Monocular, Stereo, and RGB-D Cameras}, 
  year={2017},
  volume={33},
  number={5},
  pages={1255-1262},
  doi={10.1109/TRO.2017.2705103}
}

@article{orbslam3,
  author={Campos, Carlos and Elvira, Richard and Rodríguez, Juan J. Gómez and M. Montiel, José M. and D. Tardós, Juan},
  journal=TRO, 
  title={{ORB-SLAM3}: An Accurate Open-Source Library for Visual, Visual–Inertial, and Multimap SLAM}, 
  year={2021},
  volume={37},
  number={6},
  pages={1874-1890},
  doi={10.1109/TRO.2021.3075644}
}

@inproceedings{kimera,
  author={Rosinol, Antoni and Abate, Marcus and Chang, Yun and Carlone, Luca},
  booktitle=ICRA, 
  title={Kimera: an Open-Source Library for Real-Time Metric-Semantic Localization and Mapping}, 
  year={2020},
  pages={1689-1696},
  doi={10.1109/ICRA40945.2020.9196885}
}

@inproceedings{lsdslam,
  title={{LSD-SLAM}: Large-scale direct monocular SLAM},
  author={Engel, Jakob and Sch{\"o}ps, Thomas and Cremers, Daniel},
  booktitle=ECCV,
  pages={834--849},
  year={2014},
  organization={Springer}
}

@article{dso,
  author={Engel, Jakob and Koltun, Vladlen and Cremers, Daniel},
  journal=TPAMI, 
  title={Direct Sparse Odometry}, 
  year={2018},
  volume={40},
  number={3},
  pages={611-625},
  doi={10.1109/TPAMI.2017.2658577}
}

@article{ba,
  title={Bundle Adjustment—A Modern Synthesis},
  author={Triggs, Bill and McLauchlan, Philip and Hartley, Richard and Fitzgibbon, Andrew},
  journal={Vision Algorithms: Theory and Practice},
  pages={153--177},
  year={2000},
  publisher={Springer}
}

@article{deepfactors,
  author={Czarnowski, Jan and Laidlow, Tristan and Clark, Ronald and Davison, Andrew J.},
  journal=RAL, 
  title={DeepFactors: Real-Time Probabilistic Dense Monocular SLAM}, 
  year={2020},
  volume={5},
  number={2},
  pages={721-728},
  doi={10.1109/LRA.2020.2965415}
}

@article{deepv2d,
  title={Deepv2d: Video to depth with differentiable structure from motion},
  author={Teed, Zachary and Deng, Jia},
  booktitle=ICLR,
  year={2020},
}

@article{droidslam,
  title={Droid-slam: Deep visual slam for monocular, stereo, and rgb-d cameras},
  author={Teed, Zachary and Deng, Jia},
  booktitle=NeurIPS,
  volume={34},
  pages={16558--16569},
  year={2021}
}

@inproceedings{dpvslam,
  title={Deep patch visual slam},
  author={Lipson, Lahav and Teed, Zachary and Deng, Jia},
  booktitle=ECCV,
  pages={424--440},
  year={2024},
  organization={Springer}
}

@inproceedings{imap,
  author={Sucar, Edgar and Liu, Shikun and Ortiz, Joseph and Davison, Andrew J.},
  booktitle=ICCV, 
  title={{iMAP}: Implicit Mapping and Positioning in Real-Time}, 
  year={2021},
  pages={6209-6218},
  doi={10.1109/ICCV48922.2021.00617}
}

@inproceedings{nicerslam,
  author={Zhu, Zihan and Peng, Songyou and Larsson, Viktor and Cui, Zhaopeng and Oswald, Martin R. and Geiger, Andreas and Pollefeys, Marc},
  booktitle=THREEDV, 
  title={{NICER-SLAM}: Neural Implicit Scene Encoding for RGB SLAM}, 
  year={2024},
  pages={42-52},
  doi={10.1109/3DV62453.2024.00096}
}

@article{glorieslam,
  title={Glorie-slam: Globally optimized rgb-only implicit encoding point cloud slam},
  author={Zhang, Ganlin and Sandstr{\"o}m, Erik and Zhang, Youmin and Patel, Manthan and Van Gool, Luc and Oswald, Martin R},
  journal={arXiv preprint arXiv:2403.19549},
  year={2024}
}

@article{3dgs,
      author       = {Kerbl, Bernhard and Kopanas, Georgios and Leimk{\"u}hler, Thomas and Drettakis, George},
      title        = {3D Gaussian Splatting for Real-Time Radiance Field Rendering},
      journal      = {ACM Transactions on Graphics},
      number       = {4},
      volume       = {42},
      month        = {July},
      year         = {2023},
      url          = {https://repo-sam.inria.fr/fungraph/3d-gaussian-splatting/}
}

@inproceedings{gaussiansplattingslam,
  author={Matsuki, Hidenobu and Murai, Riku and Kelly, Paul H. J. and Davison, Andrew J.},
  booktitle=CVPR, 
  title={Gaussian Splatting SLAM}, 
  year={2024},
  pages={18039-18048},
  doi={10.1109/CVPR52733.2024.01708}
}

@inproceedings{gsslam,
  author={Yan, Chi and Qu, Delin and Xu, Dan and Zhao, Bin and Wang, Zhigang and Wang, Dong and Li, Xuelong},
  booktitle=CVPR, 
  title={{GS-SLAM}: Dense Visual SLAM with 3D Gaussian Splatting}, 
  year={2024},
  pages={19595-19604},
  doi={10.1109/CVPR52733.2024.01853}
}

@article{dinov2,
    title={{DINO}v2: Learning Robust Visual Features without Supervision},
    author={Maxime Oquab and Timoth{\'e}e Darcet and Th{\'e}o Moutakanni and Huy V. Vo and Marc Szafraniec and Vasil Khalidov and Pierre Fernandez and Daniel HAZIZA and Francisco Massa and Alaaeldin El-Nouby and Mido Assran and Nicolas Ballas and Wojciech Galuba and Russell Howes and Po-Yao Huang and Shang-Wen Li and Ishan Misra and Michael Rabbat and Vasu Sharma and Gabriel Synnaeve and Hu Xu and Herve Jegou and Julien Mairal and Patrick Labatut and Armand Joulin and Piotr Bojanowski},
    journal=TMLR,
    issn={2835-8856},
    year={2024},
    url={https://openreview.net/forum?id=a68SUt6zFt},
    note={Featured Certification}
}

@inproceedings{depthanything,
  author={Yang, Lihe and Kang, Bingyi and Huang, Zilong and Xu, Xiaogang and Feng, Jiashi and Zhao, Hengshuang},
  booktitle=CVPR, 
  title={Depth Anything: Unleashing the Power of Large-Scale Unlabeled Data}, 
  year={2024},
  pages={10371-10381},
  doi={10.1109/CVPR52733.2024.00987}
}

@article{depthanythingv2,
  title={Depth anything v2},
  author={Yang, Lihe and Kang, Bingyi and Huang, Zilong and Zhao, Zhen and Xu, Xiaogang and Feng, Jiashi and Zhao, Hengshuang},
  journal=NeurIPS,
  volume={37},
  pages={21875--21911},
  year={2024}
}

@inproceedings{depthanythingv3,
    title={Depth Anything 3: Recovering the Visual Space from Any Views},
    author={Haotong Lin and Sili Chen and Jun Hao Liew and Donny Y. Chen and Zhenyu Li and Yang Zhao and Sida Peng and Hengkai Guo and Xiaowei Zhou and Guang Shi and Jiashi Feng and Bingyi Kang},
    booktitle=ICLR,
    year={2026},
    url={https://openreview.net/forum?id=yirunib8l8}
}

@inproceedings{foundationstereo,
  author={Wen, Bowen and Trepte, Matthew and Aribido, Joseph and Kautz, Jan and Gallo, Orazio and Birchfield, Stan},
  booktitle=CVPR, 
  title={FoundationStereo: Zero-Shot Stereo Matching}, 
  year={2025},
  pages={5249-5260},
  doi={10.1109/CVPR52734.2025.00495}
}

@inproceedings{dust3r,
  author={Wang, Shuzhe and Leroy, Vincent and Cabon, Yohann and Chidlovskii, Boris and Revaud, Jerome},
  booktitle=CVPR, 
  title={{DUSt3R}: Geometric 3D Vision Made Easy}, 
  year={2024},
  pages={20697-20709},
  doi={10.1109/CVPR52733.2024.01956}
}

@inproceedings{mast3r,
  title={Grounding image matching in 3d with mast3r},
  author={Leroy, Vincent and Cabon, Yohann and Revaud, J{\'e}r{\^o}me},
  booktitle=ECCV,
  pages={71--91},
  year={2024},
  organization={Springer}
}

@inproceedings{mast3rsfm,
  author={Duisterhof, Bardienus Pieter and Zust, Lojze and Weinzaepfel, Philippe and Leroy, Vincent and Cabon, Yohann and Revaud, Jerome},
  booktitle=THREEDV, 
  title={{MASt3R-SfM}: A Fully-Integrated Solution for Unconstrained Structure-from-Motion}, 
  year={2025},
  pages={1-10},
  doi={10.1109/3DV66043.2025.00008}
}

@inproceedings{reloc3r,
  author={Dong, Siyan and Wang, Shuzhe and Liu, Shaohui and Cai, Lulu and Fan, Qingnan and Kannala, Juho and Yang, Yanchao},
  booktitle=CVPR, 
  title={Reloc3r: Large-Scale Training of Relative Camera Pose Regression for Generalizable, Fast, and Accurate Visual Localization}, 
  year={2025},
  pages={16739-16752},
  doi={10.1109/CVPR52734.2025.01560}
}

@article{splatt3r,
  title={Splatt3r: Zero-shot gaussian splatting from uncalibrated image pairs},
  author={Smart, Brandon and Zheng, Chuanxia and Laina, Iro and Prisacariu, Victor Adrian},
  journal={arXiv preprint arXiv:2408.13912},
  year={2024}
}

@inproceedings{spann3r,
  author={Wang, Hengyi and Agapito, Lourdes},
  booktitle=THREEDV, 
  title={3D Reconstruction with Spatial Memory}, 
  year={2025},
  pages={78-89},
  doi={10.1109/3DV66043.2025.00013}
}

@inproceedings{cotracker,
  title={Cotracker: It is better to track together},
  author={Karaev, Nikita and Rocco, Ignacio and Graham, Benjamin and Neverova, Natalia and Vedaldi, Andrea and Rupprecht, Christian},
  booktitle=ECCV,
  pages={18--35},
  year={2024},
  organization={Springer}
}

@inproceedings{cut3r,
  author={Wang, Qianqian and Zhang, Yifei and Holynski, Aleksander and Efros, Alexei A. and Kanazawa, Angjoo},
  booktitle=CVPR, 
  title={Continuous 3D Perception Model with Persistent State}, 
  year={2025},
  pages={10510-10522},
  doi={10.1109/CVPR52734.2025.00983}
}

@inproceedings{vggt,
  author={Wang, Jianyuan and Chen, Minghao and Karaev, Nikita and Vedaldi, Andrea and Rupprecht, Christian and Novotny, David},
  booktitle=CVPR, 
  title={{VGGT}: Visual Geometry Grounded Transformer}, 
  year={2025},
  pages={5294-5306},
  doi={10.1109/CVPR52734.2025.00499}
}

@inproceedings{pi3,
    title={{$\pi^3$}: Permutation-Equivariant Visual Geometry Learning},
    author={Yifan Wang and Jianjun Zhou and Haoyi Zhu and Wenzheng Chang and Yang Zhou and Zizun Li and Junyi Chen and Jiangmiao Pang and Chunhua Shen and Tong He},
    booktitle=ICLR,
    year={2026},
    url={https://openreview.net/forum?id=DTQIjngDta}
}

@article{vggtlong,
  title={{VGGT-Long}: Chunk it, Loop it, Align it--Pushing VGGT's Limits on Kilometer-scale Long RGB Sequences},
  author={Deng, Kai and Ti, Zexin and Xu, Jiawei and Yang, Jian and Xie, Jin},
  journal={arXiv preprint arXiv:2507.16443},
  year={2025}
}

@inproceedings{vistaslam,
  author    = {Zhang, Ganlin and Qian, Shenhan and Wang, Xi and Cremers, Daniel},
  booktitle = THREEDV,
  title     = {{ViSTA-SLAM}: Visual {SLAM} with Symmetric Two-View Association},
  year      = {2026},
  pages     = {396-406},
  doi       = {10.1109/3DV69130.2026.00044}
}

@inproceedings{mast3rslam,
  author={Murai, Riku and Dexheimer, Eric and Davison, Andrew J.},
  booktitle=CVPR, 
  title={{MASt3R-SLAM}: Real-Time Dense SLAM with 3D Reconstruction Priors}, 
  year={2025},
  pages={16695-16705},
  doi={10.1109/CVPR52734.2025.01556}
}

@article{vggtslam,
  title={{VGGT-SLAM}: Dense RGB SLAM Optimized on the SL (4) Manifold},
  author={Maggio, Dominic and Lim, Hyungtae and Carlone, Luca},
  journal=NeurIPS,
  volume={38},
  year={2025}
}

@inproceedings{goslam,
  author={Zhang, Youmin and Tosi, Fabio and Mattoccia, Stefano and Poggi, Matteo},
  booktitle=ICCV, 
  title={{GO-SLAM}: Global {O}ptimization for {C}onsistent 3D {I}nstant {R}econstruction}, 
  year={2023},
  pages={3704-3714},
  doi={10.1109/ICCV51070.2023.00345}
}

@inproceedings{slam3r,
  author={Liu, Yuzheng and Dong, Siyan and Wang, Shuzhe and Yin, Yingda and Yang, Yanchao and Fan, Qingnan and Chen, Baoquan},
  booktitle=CVPR, 
  title={{SLAM3R}: Real-Time Dense Scene Reconstruction from Monocular RGB Videos}, 
  year={2025},
  pages={16651-16662},
  doi={10.1109/CVPR52734.2025.01552}
}

@inproceedings{brachmann2021limits,
  author={Brachmann, Eric and Humenberger, Martin and Rother, Carsten and Sattler, Torsten},
  booktitle=ICCV, 
  title={On the Limits of Pseudo Ground Truth in Visual Camera Re-localisation}, 
  year={2021},
  pages={6198-6208},
  doi={10.1109/ICCV48922.2021.00616}
}

@misc{evo,
  title={evo: Python package for the evaluation of odometry and SLAM.},
  author={Grupp, Michael},
  howpublished={\url{https://github.com/MichaelGrupp/evo}},
  year={2017}
}

@article{replica,
  title =   {The {R}eplica Dataset: A Digital Replica of Indoor Spaces},
  author =  {Julian Straub and Thomas Whelan and Lingni Ma and Yufan Chen and Erik Wijmans and Simon Green and Jakob J. Engel and Raul Mur-Artal and Carl Ren and Shobhit Verma and Anton Clarkson and Mingfei Yan and Brian Budge and Yajie Yan and Xiaqing Pan and June Yon and Yuyang Zou and Kimberly Leon and Nigel Carter and Jesus Briales and  Tyler Gillingham and  Elias Mueggler and Luis Pesqueira and Manolis Savva and Dhruv Batra and Hauke M. Strasdat and Renzo De Nardi and Michael Goesele and Steven Lovegrove and Richard Newcombe },
  journal = {arXiv preprint arXiv:1906.05797},
  year =    {2019}
}

@InProceedings{salad,
    author    = {Izquierdo, Sergio and Civera, Javier},
    title     = {{O}ptimal {T}ransport {A}ggregation for {V}isual {P}lace {R}ecognition},
    booktitle = {Proceedings of the IEEE/CVF Conference on Computer Vision and Pattern Recognition (CVPR)},
    month     = {June},
    year      = {2024},
}

@inproceedings{mapanything,
  title={Mapanything: Universal feed-forward metric 3d reconstruction; map-anything. github. io},
  author={Keetha, Nikhil and M{\"u}ller, Norman and Sch{\"o}nberger, Johannes and Porzi, Lorenzo and Zhang, Yuchen and Fischer, Tobias and Knapitsch, Arno and Zauss, Duncan and Weber, Ethan and Antunes, Nelson and others},
  booktitle={2026 International Conference on 3D Vision (3DV)},
  pages={499--509},
  year={2026},
  organization={IEEE}
}
